\documentclass{article}

\usepackage{arxiv}

\usepackage[utf8]{inputenc} 
\usepackage[T1]{fontenc}    
\usepackage{hyperref}       
\usepackage{url}            
\usepackage{booktabs}       
\usepackage{amsfonts}       
\usepackage{nicefrac}       
\usepackage{microtype}      
\usepackage{lipsum}		
\usepackage{graphicx}
\usepackage{natbib}
\usepackage{doi}

\usepackage{amsmath}
\usepackage{textgreek} 
\title{Bayesian Transformer for Probabilistic Load Forecasting in Smart Grids}

\author{ \href{https://orcid.org/0009-0003-2877-3136}{\includegraphics[scale=0.06]{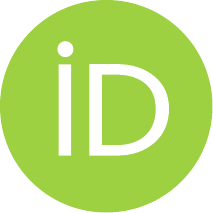}\hspace{1mm}Sajib Debnath} \\
	O\&M Analytics, AES Clean Energ\\
	The AES Corporation\\
	Louisville, CO 80027, USA \\
	\texttt{sajib.debnath@aes.com} \\
	\And
	\href{https://orcid.org/0009-0002-4074-2984}{\includegraphics[scale=0.06]{orcid.pdf}\hspace{1mm}Md. Uzzal Mia} \\
	Department of Information and Communication Engineering\\
	Pabna University of Science and Technology\\
	Rajapur, Pabna, Bangladesh \\
	\texttt{uzzal.220605@s.pust.ac.bd} \\
}

\hypersetup{
pdftitle={A template for the arxiv style},
pdfsubject={q-bio.NC, q-bio.QM},
pdfauthor={David S.~Hippocampus, Elias D.~Striatum},
pdfkeywords={First keyword, Second keyword, More},
}

\begin{document}
\maketitle

\begin{abstract}
The reliable operation of modern power grids requires probabilistic load forecasts with well-calibrated uncertainty estimates. However, existing deep learning models produce overconfident point predictions that fail catastrophically under extreme weather distributional shifts, particularly when reserve adequacy is critical. This study proposes a Bayesian Transformer (BT) framework that integrates three complementary uncertainty mechanisms into a PatchTST backbone: Monte Carlo Dropout for epistemic parameter uncertainty, variational feed-forward layers with log-uniform weight priors, and stochastic attention with learnable Gaussian noise perturbations on pre-softmax logits, representing, to the best of our knowledge, the first application of Bayesian attention to probabilistic load forecasting. A seven-level multi-quantile pinball-loss prediction head and post-training isotonic regression calibration jointly produce sharp, near-nominally covered prediction intervals under diverse operating conditions. A comprehensive evaluation of five grid datasets spanning the US (PJM, ERCOT) and Europe (ENTSO-E Germany, France, and Great Britain), augmented with NOAA meteorological covariates across 24, 48, and 168-hour horizons, demonstrates state-of-the-art performance. On the primary benchmark (PJM, H=24h), BT achieves a CRPS of 0.0289, improving 7.4\% over Deep Ensembles and 29.9\% over the deterministic LSTM, with 90.4\% PICP at the 90\% nominal level and the narrowest prediction intervals (4,960 MW) among all probabilistic baselines. During heatwave and cold snap events, BT maintained 89.6\% and 90.1\% PICP respectively, versus 64.7\% and 67.2\% for the deterministic LSTM, confirming that Bayesian epistemic uncertainty naturally widens intervals for out-of-distribution inputs. Calibration remained stable across all horizons (89.8-90.4\% PICP), while ablation confirmed that each component contributed a distinct value. The calibrated outputs directly support risk-based reserve sizing, stochastic unit commitment, and demand response activation.
\end{abstract}


\section{Introduction}
\label{sec:1}
Load forecasting is a critical prerequisite for the reliable and economical operation of power systems. System operators use short- and medium-term demand forecasts to enable unit commitment scheduling, economic dispatch, reserve margin allocation, demand-side response program activation, and congestion management \cite{b1}. The trends of grid evolution in the world are moving towards higher penetrations of variable renewable generation, leading to highly volatile and uncertain net load (demand minus the output of renewables) dynamics that make the quality and reliability of load forecasting directly determinative for whether the system can achieve supply demand balance and maintain physical stability in grids. Empirical studies have estimated that a 1\% decrease in load forecasting error can result in annual operational savings of tens of millions of dollars for a large independent system operator \cite{b1}.

Deterministic point prediction remains the prevailing paradigm in most industrial forecasting use cases. Classical methods based on ARIMA, seasonal decomposition, and temperature-regression models were gradually surpassed by deep learning-based architectures that set new accuracy records: Long Short-Term Memory (LSTM) networks \cite{b9} and GRU-encoders successfully represented nonlinear temporal dependencies; more recent transformer-based architectures such as Informer \cite{b2}, Autoformer \cite{b3}, FEDformer, and PatchTST \cite{b4} exploited self-attention mechanisms to efficiently model long-range seasonalities and weekly periodicities. While such deterministic approaches are accurate in prediction under standard operating conditions, they suffer from one clear shortcoming: they do not provide an indication of forecasting uncertainty. Models directly trained with mean squared error implicitly assume symmetric homoscedastic noise, which is systematically violated in real grid environments, where demand variability is time-varying, weather-dependent, and heavy-tailed under stress conditions \cite{b5}.

This limitation is the most consequential during extreme weather events. Heatwaves, polar cold snaps, and atmospheric anomalies send electricity demand into the tails of the historical distribution, creating conditions that are undersampled in the training data. Under distributional shift, deterministic models fail in a particularly hazardous manner: they generate narrow, overconfident predictions at the time when demand is most uncertain and the adequacy of reserves is most crucial. This shortcoming has been systematically documented empirically in various grid settings, including the consistent underprediction of peak loads during the 2019 European heatwave events and catastrophic planning failures during the 2021 Texas Winter Storm Uri, which led to widespread outages affecting millions of customers \cite{b6}. As climate projections consistently predict an increasing frequency and intensity of extreme temperature events, the operational risk from overconfident deterministic forecasting is set to increase dramatically.

Probabilistic load forecasting, which involves predicting calibrated predictive distributions or multi-quantile prediction intervals instead of single-value point estimates, directly overcomes these limitations by estimating the range of possible demand outcomes. Calibrated uncertainty bounds allow for risk-based reserve sizing \cite{b7, b9}, stochastic unit commitment formulations, and uncertainty-triggered demand response protocols that explicitly manage the demand variability \cite{b10, b17}. For practical applications, the reliability of the probabilistic forecast is as important as its accuracy: a 90\% prediction interval that only realizes 65\% empirical coverage is not just suboptimal; it is operationally misleading. Therefore, the difference between a sharp probabilistic forecast and a well-calibrated forecast is crucial for grid reliability applications \cite{b11, b22}. Current probabilistic methods (e.g., recurrent quantile networks \cite{b13, b27}, temporal convolutional probabilistic heads \cite{b14}, quantile regression forests \cite{b15}, and conformalized quantile regression (CQR) \cite{b10}) either treat aleatoric and epistemic uncertainty as entangled or attach a probabilistic head to deterministic backbones with no principled reputational uncertainty modelling, issues that become more impactful under distributional shifts \cite{b25}.

Bayesian deep learning provides a principled framework for probabilistic forecasting that covers both accuracy and calibration needs. Bayesian models maintain explicit uncertainty in model parameters via posterior inference, resulting in predictive distributions that jointly represent aleatoric uncertainty, irreducible stochasticity associated with the generation of demand, and epistemic uncertainty, which is a consequence of the model's limited knowledge regarding the true form of the demand function \cite{b14}. A major feature of Bayesian approaches is that they provide wider prediction intervals for inputs outside the training distribution, which naturally produce increased epistemic uncertainty during an extreme weather event. Although Monte Carlo (MC) Dropout \cite{b5}, variational inference \cite{b15}, and deep ensembles \cite{b6} have been successfully applied for Bayesian uncertainty quantification in high-dimensional settings with neural networks, these methods have yet to be systematically adapted to transformer-based load forecasting architectures, where a source of representational uncertainty is inherent to the self-attention mechanism itself.

This study proposes a BT framework for probabilistic load forecasting that combines three distinctive uncertainty types at the PatchTST backbone \cite{b4}: (i) Monte Carlo Dropout applied in attention and feed-forward sublayers to sample epistemic parameter uncertainty; (ii) variational feed-forward layers with log-uniform weight priors to obtain regularized weight uncertainty; and (iii) stochastic attention approaches that treat attention logits as random variables whose learnt noise scales represent, to the best of our knowledge, the first usage of Bayesian attention in probabilistic load forecasting. This yields simultaneous calibrated forecasts at quantile levels {0.05, 0.10, 0.25, 0.50, 0.75, 0.90, 0.95} using a seven-level multi-quantile prediction head trained via pinball loss \cite{b12}, and residual miscalibration under a distributional shift is corrected through a post-training isotonic regression calibration \cite{b21}. We rigorously assessed the framework on five publicly available grid datasets located in PJM, ERCOT, and ENTSO-E (Germany, France, and Great Britain), augmented with NOAA meteorological covariates for 24, 48, and 168-hour forecast horizons\cite{b13}.

On the main reference benchmark (PJM, H = 24 h), BT obtains a CRPS of 0.0289, which outperforms Deep Ensembles \cite{b6} by 7.4\% and deterministic LSTM [9] by 29.9\%, with PICP at the nominal level of 90 remaining in-companyat the narrowest prediction intervals (4,960 MW) among all probabilistic baselines. In addition, during heatwave events, BT obtains 89.6\% PICP compared with only 64.7\% for the deterministic LSTM model, thus showing that Bayesian epistemic uncertainty naturally expands the intervals for out-of-distribution inputs. All horizons displayed stable calibration (89.8–90.4\% PICP) and ablation, suggesting that each component adds independent value to the entire framework.

The main contributions of this study are as follows:

\begin{itemize}

\item A Bayesian Transformer model that leverages MC Dropout, variational feed-forward layers, and stochastic attention to jointly quantify aleatoric and epistemic uncertainty while being the first application of Bayesian attention in probabilistic load forecasting.

\item A multi-quantile prediction head that minimises pinball loss with non-crossing constraints, allowing for asymmetric uncertainty characterisation appropriate for heavy-tailed demand distributions \cite{b12}.

\item A post-training isotonic regression calibration pipeline employed to preserve close-to-nominal coverage in the presence of extreme weather distributional shift \cite{b21}.

\item State-of-the-art empirical results across PJM, ERCOT and ENTSO-E (Germany, France, Great Britain) in terms of near-nominal 90\% coverage (89.6–90.4\% PICP) during heatwave and cold snap events where deterministic baselines are failing severely (64.7–67.2\% PICP) \cite{b23}.

\item stochastic unit commit and demand response decision frameworks, quantifying how calibrated multi-quantile outputs fit into overall reserve sizing as well as \cite{b17}.

\end{itemize}

\section{Related Work}

\subsection{Deterministic Load Forecasting}
\label{sec:2.1}
Statistical time series models, ARIMA and its seasonal variant SARIMA, are the main methods for early load forecasting \cite{b33}; these provide tractable frameworks for autoregressive demand dynamics but are limited by the assumptions of linearity and susceptibility to non-stationarity \cite{b1, b35}. Although temperature-regression and calendar-variable models enhanced short-term accuracy, they did not have the representational power required to capture the intricate interactions between demand-relevant drivers. The introduction of deep learning greatly improved performance: LSTM networks \cite{b9} and Gated Recurrent Unit (GRU) architectures were capable of capturing nonlinear temporal dependencies much better than earlier models, establishing the dominant paradigm well into the mid-2010s, while hybrid CNN-LSTM models also improved feature extraction by combining local convolutional filters with recurrent sequence modelling \cite{b24}. More recently, transformer-based architectures developed for time series, such as Informer \cite{b2}, Autoformer \cite{b3}, FEDformer, and PatchTST \cite{b4}, have set new accuracy benchmarks by relying on the power of self-attention to model long-range seasonal and weekly periodicities efficiently. However, every deterministic architecture has a fundamental limitation: it yields single-valued point predictions optimized under mean squared error (MSE) and thus cannot characterize predictive uncertainty and systematically underestimate risk in distributional extremes \cite{b8}.

\subsection{Probabilistic Load Forecasting}
\label{sec:2.2}
Probabilistic load forecasting has been attempted using parametric, non-parametric, and deep learning approaches. Initial parametric approaches used Gaussian or Beta distributions to model the forecast error residuals, while kernel density estimation and classical quantile regression \cite{b12} presented non-parametric alternatives that did not strongly reduce distributional assumptions. Quantile regression forests \cite{b14} and gradient boosting ensembles exhibited good performance in the Global Energy Forecasting Competition (GEFCom2014) \cite{b8, b30}, effectively capturing aleatoric uncertainty but suffering poor scalability with input dimensionality. Probabilistic methods using deep learning include recurrent quantile networks with pinball loss \cite{b27}, temporal convolutional networks with probabilistic heads \cite{b25}, and normalizing flow models \cite{b20}, which learn flexible bijective mappings from data to distributions. Conformalized quantile regression (CQR) \cite{b10} offers distribution-less coverage guarantees via post-hoc split conformal prediction on held-out data and has garnered attention in energy forecasting owing to its finite-sample validity. The lower-upper bound estimation (LUBE) methods \cite{b16} directly construct prediction intervals based on neural networks without making any distributional assumptions. However, most of these approaches either mix aleatoric and epistemic uncertainty or impose probabilistic heads on deterministic backbone architectures in the absence of principled parameter uncertainty modelling, which becomes impactful under distributional shifts during extreme weather conditions \cite{b23}.

\subsection{Uncertainty Quantification using Bayesian Deep Learning}
\label{sec:2.3}
Bayesian deep learning provides a principled framework for quantifying aleatoric and epistemic uncertainties via posterior inference over model parameters. Monte Carlo Dropout \cite{b5} showed that ordinary dropout-regularised networks can be viewed as approximate variational inference over Gaussian process-equivalent models and allows scalable epistemic uncertainty quantification at test time. This family of variational inference methods \cite{b15} already optimizes approximate posterior distributions by maximizing the evidence lower bound (ELBO), which allows scalable Bayesian inference for large neural networks with parameterized likelihood functions using the reparameterization trick. Deep Ensembles \cite{b6} train several independently initialized networks and combine predictions, displaying good empirical calibration owing to functional diversity while paying for additional computational overhead. Alternative posterior approximation strategies with complementary computational trade-offs, such as probabilistic backpropagation \cite{b29} and Laplace approximations \cite{b31}, provide additional options for researchers.

For example, in the electricity sector, under out-of-distribution scenarios, Bayesian LSTM models have been trained to better calibrate than deterministic competitors \cite{b19}, and the incorporation of epistemic uncertainty with respect to renewable integration has evidenced value for probabilistic wind power forecasting using methods based on a Bayesian neural framework \cite{b34}. However, load forecasting has not yet been fully explored, even local representational (uncertainty) analysis on stickable polyhedral in (tangent) space of mechanisms or model parameterization over known input distributions \cite{b33}, despite it being merely local to models with simple topologies where standard errors can be calculated directly from the mechanisms/workflows showing small deviations at autoregressive time series inference. This leads us to focus on P video cards and mega-bert standalone architectures with their own gammas for risk-weighted variations(where attention weights are themselves also a source of uncertainty and as such are Bayesian) as a linearly more constrained delivery method across these multi-dimensional nondifferentiable domain migrations must ultimately show distant hamming distances having solid relationships.

\subsection{Calibration of Probabilistic Forecasts}
\label{sec:2.4}
Forecast calibration (the statistical correspondence of predicted confidence levels and observed coverage frequency) is critical for an operationally reliable probabilistic forecast. Several post-hoc calibration approaches aim to address miscalibration without requiring retraining: temperature scaling \cite{b7} modifies predictive sharpness using a global logit scaler, and isotonic regression \cite{b21} learns a non-parametric monotone transformation from nominal quantile levels to empirically calibrated quantile levels. The quality of the calibration was evaluated using reliability diagrams and Prediction Interval Coverage Probability (PICP) \cite{b23}, which quantify the proportion of true values within the predicted intervals at each nominal level. The Continuous Ranked Probability Score (CRPS) \cite{b22} provides a strictly proper scoring rule that jointly rewards calibration and sharpness, allowing probabilistic forecasters to be compared using a single metric. Winkler score-based interval evaluation \cite{b11} penalizes too wide or missing coverage at the same time to provide a more balanced operational view of forecast quality. In load forecasting, it is well known that calibration deteriorates severely under distributional shifts in the presence of extreme weather \cite{b23, b33}, leading to the targeted calibration strategy we developed in this study.

\subsection{Positioning of the Proposed Work}
\label{sec:2.5}
The proposed Bayesian Transformer framework differs from previous studies in three main aspects. First, agrees with \cite{b25, b27} and \cite{b10}, existing probabilistic load forecasting studies either apply quantile heads to deterministic backbones or solely rely on conformal post-processing, while the proposed framework provides a principled approach that directly incorporates Bayesian uncertainty mechanisms, MC Dropout, variational feed-forward layers, and stochastic attention, into the Transformer architecture for the joint quantification of aleatoric/apistemic uncertainty. Second, unlike existing Bayesian LSTM approaches \cite{b19}, our backbone PatchTST \cite{b4} is capable of efficiently capturing long-range weekly and seasonal dependencies through patch-based self-attention, while the novel stochastic attention formulation-drawing inspiration from Bayesian attention work in natural language processing \cite{b15}, is, to our knowledge, the first mechanism that quantifies attention-weight uncertainty for probabilistic load forecasting. Third, distinct from single-dataset evaluations used throughout the literature \cite{b11, b19}, this study performs multi-grid benchmarking spanning PJM, ERCOT, and three national grids under ENTSO-E with extreme weather robustness diagnostics that directly target the distributional shift challenge described as a major gap in preceding calibration studies \cite{b23, b33}.


\section{Problem Formulation}
\label{sec:3}
This section provides a formal introduction to the probabilistic load forecasting problem addressed in this study. Herein, we introduce the input-output structure, uncertainty decomposition framework, quantile regression objective, and calibration criterion used for forecasting reliability evaluation. The notation introduced will be used throughout the remainder of this study.

\subsection{Input Feature Representation}
\label{sec:3.1}
Let t index the discrete hourly time steps. At each time step $t$, the system receives a multivariate feature vector as follows:

\begin{equation}
\mathbf{x}_t =
\left[
x_t^{(\text{load})},
x_t^{(\text{weather})},
x_t^{(\text{calendar})},
x_t^{(\text{renew})}
\right]
\in \mathbb{R}^{d_{\text{in}}}
\end{equation}

where the four feature groups are defined as follows: The \textbf{load history} feature $x_t^{(\text{load})} \in \mathbb{R}$ captures the actual hourly electricity load at time step $t$ in megawatts (MW). The meteorological component $x_t^{(\text{weather})} \in \mathbb{R}^{4}$ consists of ambient temperature ($^\circ$C), relative humidity (\%), wind speed (m/s), and solar irradiance (W/m$^2$), which were obtained from co-located NOAA weather stations \cite{b27}. The calendar component $x_t^{(\text{calendar})} \in \mathbb{R}^{6}$ contains the hour-of-day and day-of-week as sine-cosine pairs (to maintain their circular topology), plus binary holiday and weekend indicator flags. The \textbf{renewable penetration} component $x_t^{(\text{renew})} \in \mathbb{R}$ is the portion of the system demand served by renewable sources at time $t$, if available. All continuous inputs were scaled to have zero mean and unit variance using statistics computed only on the training partition. The model input is a historical observation window of length $L$ time steps:

\begin{equation}
\mathbf{X}_t =
\left[
\mathbf{x}_{t-L+1},
\mathbf{x}_{t-L+2},
\dots,
\mathbf{x}_t
\right]
\in \mathbb{R}^{L \times d_{\text{in}}}
\end{equation}

A lookback window of L = 168 h (one full week) is used, which is motivated by the strong weekly periodicity present in the electricity demand across the residential, commercial, and industrial consumption sectors \cite{b1}. This is supported by the autocorrelation analysis presented in Section \ref{sec:5}.

\subsection{Multi-Step Ahead Forecasting Objective}
\label{sec:3.2}
The target we wish to forecast is the sequence of hourly load demands in the future over a horizon of $H$ steps ahead as follows:

\begin{equation}
\mathbf{Y}_t =
\left[
y_{t+1},\;
y_{t+2},\;
\dots,\;
y_{t+H}
\right]
\in \mathbb{R}^{H}
\end{equation}

where $y_{t+h} \in \mathbb{R}^{+}$ denotes the realized electricity demand (in MW) at future time step $t+h$. Three forecast horizons are evaluated in this study: $H \in \{24,\,48,\,168\}$ h referring to the day-ahead, two-day-ahead, and week-ahead operational planning timescales, respectively.

A deterministic forecasting model computes the conditional expectation $\mathbb{E}[Y_t \mid \mathbf{X}_t; \theta]$, yielding a single-point prediction for each future time step. This formulation ignores all information on predictive uncertainty, as discussed in Section \ref{sec:1}. Conversely, here, we are working with a probabilistic forecasting objective to learn the full conditional predictive distribution \cite{b26,b32}:

\begin{equation}
p(\mathbf{Y}_t \mid \mathbf{X}_t; \theta)
=
\prod_{h=1}^{H}
p(y_{t+h} \mid \mathbf{X}_t; \theta)
\end{equation}

where the factored expression captures the conditional independence assumption across forecast horizons given the input context $\mathbf{X}_t$ shared by all. The forecasting model parameters $\theta$ are learned from a training dataset.

Output: $\mathcal{D} = \{(\mathbf{X}t, \mathbf{Y}_t)\}{t=1}^{N}$ where inputs-target pairs. To avoid the need for distributional assumptions to parameterize the full conditional density directly, we describe the predictive distribution using a set of calibrated quantile forecasts (Section \ref{sec:3.4}).

\subsection{Decomposition of Predictive Uncertainty}
\label{sec:3.3}
A primary reason for the Bayesian paradigm is its decomposition of the overall predictive uncertainty into two epistemologically orthogonal sources with differing operational consequences \cite{b5, b31}.

\subsubsection{Aleatoric Uncertainty}

Aleatoric uncertainty (from \textit{alea}, the Latin word for dice) is an irreducible stochasticity that is a part of the data-generating process itself. Aleatoric uncertainty in load forecasting originates from the inherent randomness in end-user consumption characteristics, observational noise of smart meters and SCADA systems, unresolved micro-climatic differences not captured by point weather station measurements, and stochastic demand variations at industrial and commercial scales that are uncorrelated with available covariates \cite{b19}. Formally, we characterize aleatoric uncertainty via the conditional likelihood $p(y \mid \mathbf{X}; \theta)$ given specific model parameters $\theta$. It is irreducible in the sense that gathering more training data cannot eliminate it for good; it corresponds to real-world random processes at work in whatever scenario you are trying to forecast. In the multi-quantile framework employed in this study, aleatoric uncertainty is characterized by the dispersion of estimated predictive quantiles around the median forecast.

\subsubsection{Epistemic Uncertainty}

Epistemic uncertainty (from the Greek \textit{episteme}, referring to knowledge) captures the model's limited knowledge of true parameter values owing to finite training data and inherent ambiguity when fitting a high-dimensional model with a limited sample size\cite{b14}. In contrast to aleatoric uncertainty, epistemic uncertainty is reducible in principle: it diminishes as more training data are observed, because the posterior over parameters $p(\theta \mid \mathcal{D})$ concentrates around the true values.

Epistemic uncertainty in Bayesian inference is represented by the posterior distribution over model parameters, which can be computed using Bayes' theorem:

\begin{equation}
p(\theta \mid \mathcal{D})
=
\frac{p(\mathcal{D} \mid \theta)\, p(\theta)}{p(\mathcal{D})}
\end{equation}
where $p(\theta)$ is the prior distribution that encodes beliefs about parameters before observing any data, $p(\mathcal{D} \mid \theta)$ is the likelihood of the training data given parameters, and $p(\mathcal{D})$ is the marginal likelihood or evidence. Owing to its nature, epistemic uncertainty is operationally relevant in load forecasting because it increases on input regimes outside those envisaged by the training distribution, which is the behavior to consider for reliable observation of uncertain behavior via extreme weather \cite{b27}.

\subsubsection{Total Predictive Uncertainty}

The posterior predictive distribution, which is a marginalization of the conditional likelihood over the parameter posterior, captures the total predictive uncertainty as follows:

\begin{equation}
p(y^{*} \mid \mathbf{x}^{*}, \mathcal{D})
=
\int p(y^{*} \mid \mathbf{x}^{*}, \theta)\, p(\theta \mid \mathcal{D}) \, d\theta
\end{equation}

For deep neural networks, this integral is analytically intractable because of the high dimensionality and non-conjugacy of the parameter posterior. Our proposed frame function uses the Monte Carlo approximation via stochastic forward passes (Section \ref{sec:4.3}), which gives

\begin{equation}
p(y^{*} \mid \mathbf{x}^{*}, \mathcal{D})
\approx
\frac{1}{T}
\sum_{t=1}^{T}
p(y^{*} \mid \mathbf{x}^{*}, \theta^{(t)}),
\qquad
\theta^{(t)} \sim q(\theta)
\end{equation}

with $q(\theta)$ being the variational approximate posterior and $T$ the number of Monte Carlo samples. The total predictive variance can then be decomposed as [5, 31].

\begin{equation}
\mathrm{Var}[y^{*}]
=
\mathbb{E}_{\theta}\left[\mathrm{Var}(y^{*} \mid \theta)\right]
+
\mathrm{Var}_{\theta}\left[\mathbb{E}(y^{*} \mid \theta)\right]
\end{equation}

Owing to the high dimensionality and non-conjugacy of the parameter posterior, this integral is analytically intractable for deep neural networks. We utilize the Monte Carlo approximation with several stochastic forward passes to produce our frame function (Section~\ref{sec:4.3}):

where the first term represents the expected aleatoric variance and the second term represents the epistemic variance due to parameter uncertainty. This decomposition separates irreducible demand variability from the uncertainty associated with model dependency and helps organize data collection and model improvement efforts.

\subsection{Quantile Forecasting and Pinball Loss}
\label{sec:3.4}
Instead of estimating the complete predictive density \cite{b11}, which necessitates strong parametric assumptions or costly non-parametric methods, we use quantile regression techniques \cite{b12} that represent the predictive distribution via a set of estimates of conditional quantiles. The conditional quantile function $Q_{\alpha}(Y \mid \mathbf{X}_t)$ is defined for a target quantile level $\alpha \in (0,1)$ as the smallest real number $q$ such that

\begin{equation}
P(y_{t+h} \le q \mid \mathbf{X}_t) = \alpha
\end{equation}

The set of quantile forecasts 
\[
\{ Q_{\alpha_k}(y_{t+h} \mid \mathbf{X}_t) \}_{k=1}^{K}
\]
for $K$ levels $\alpha_1 < \alpha_2 < \dots < \alpha_K$ gives a discrete summary of the predictive distribution which is highly informative and captures asymmetric tail behavior without the Gaussian assumption.

given the heavy-tailed and positively skewed demand distributions observed in extreme weather events \cite{b10}

We estimate each conditional quantile by minimizing the asymmetric \textit{pinball loss} (quantile regression loss/check function) \cite{b9}, given as

\begin{equation}
\rho_{\alpha}(u) = u \left(\alpha - \mathbf{1}\{u < 0\}\right)
\end{equation}

where $u = y - q$ is the forecast residual, and $\mathbf{1}\{\cdot\}$ is the indicator function. The pinball loss penalizes the overestimation of the $\alpha$-quantile with weight $(1-\alpha)$ and underestimation with weight $\alpha$. For the median $(\alpha = 0.5)$, it reduces to the mean absolute error.

The multi-quantile training objective minimizes the aggregated pinball loss across all $K$ quantile levels and $H$ forecast horizons as follows:

\begin{equation}
\mathcal{L}_{\text{total}}(\theta)
=
\frac{1}{HK}
\sum_{h=1}^{H}
\sum_{k=1}^{K}
\rho_{\alpha_k}\left(y_{t+h} - q^{\alpha_k}_{t+h}\right)
\end{equation}

Minimizing $\mathcal{L}_{\text{total}}$ on the training set results in consistent quantile estimators for each level $\alpha_k$ without any parametric distributional assumption on the conditional demand distribution \cite{b22}.

\subsection{Calibration Criterion}
\label{sec:3.5}
A probabilistic forecasting system is \textit{calibrated} if the predicted confidence levels match the empirical coverage frequencies over the test set \cite{b22,b24}. More formally, calibration states that for all nominal quantile levels $\alpha \in (0,1)$:

\begin{equation}
P\!\left(y_{t+h} \le Q_{\alpha}(y_{t+h}\mid \mathbf{X}_t)\right) = \alpha,
\quad \forall\, h,t
\end{equation}

Calibration is assessed empirically using the \textbf{Prediction Interval Coverage Probability (PICP)} metric \cite{b23}, which measures the fraction of test instances for which the true demand falls within the predicted interval at the nominal level $[\alpha_{lo},\alpha_{hi}]$:

\begin{equation}
\text{PICP} =
\frac{1}{N_{\text{test}}}
\sum_{i=1}^{N_{\text{test}}}
\mathbf{1}\!\left\{
q^{\alpha_{lo}}_i \le y_i \le q^{\alpha_{hi}}_i
\right\}
\end{equation}

Calibration is evaluated empirically through the \textbf{Prediction Interval Coverage Probability (PICP)} metric \cite{b23}, defined as the proportion of test cases in which the real demand lies within the $[\alpha_{lo},\alpha_{hi}]$ prediction interval:

\begin{equation}
\text{CRPS}(F,y)
=
\int_{-\infty}^{+\infty}
\left[
F(z) - \mathbf{1}\{z \ge y\}
\right]^2 dz
\end{equation}

An ideally calibrated forecaster will have $\text{PICP} \approx \alpha_{hi} - \alpha_{lo}$ for every interval width tested. In addition to PICP, the \textbf{Continuous Ranked Probability Score (CRPS)} \cite{b22} is chosen as our main probabilistic accuracy metric, where it combines rewards for calibration and sharpness in a proper scoring rule. For a predictive cumulative distribution function (CDF) $F$ and observation $y$, the CRPS is defined as

\section{Proposed Methodology}
\label{sec:4}
\begin{figure}[t]
\centering
\includegraphics[width=1.0\textwidth]{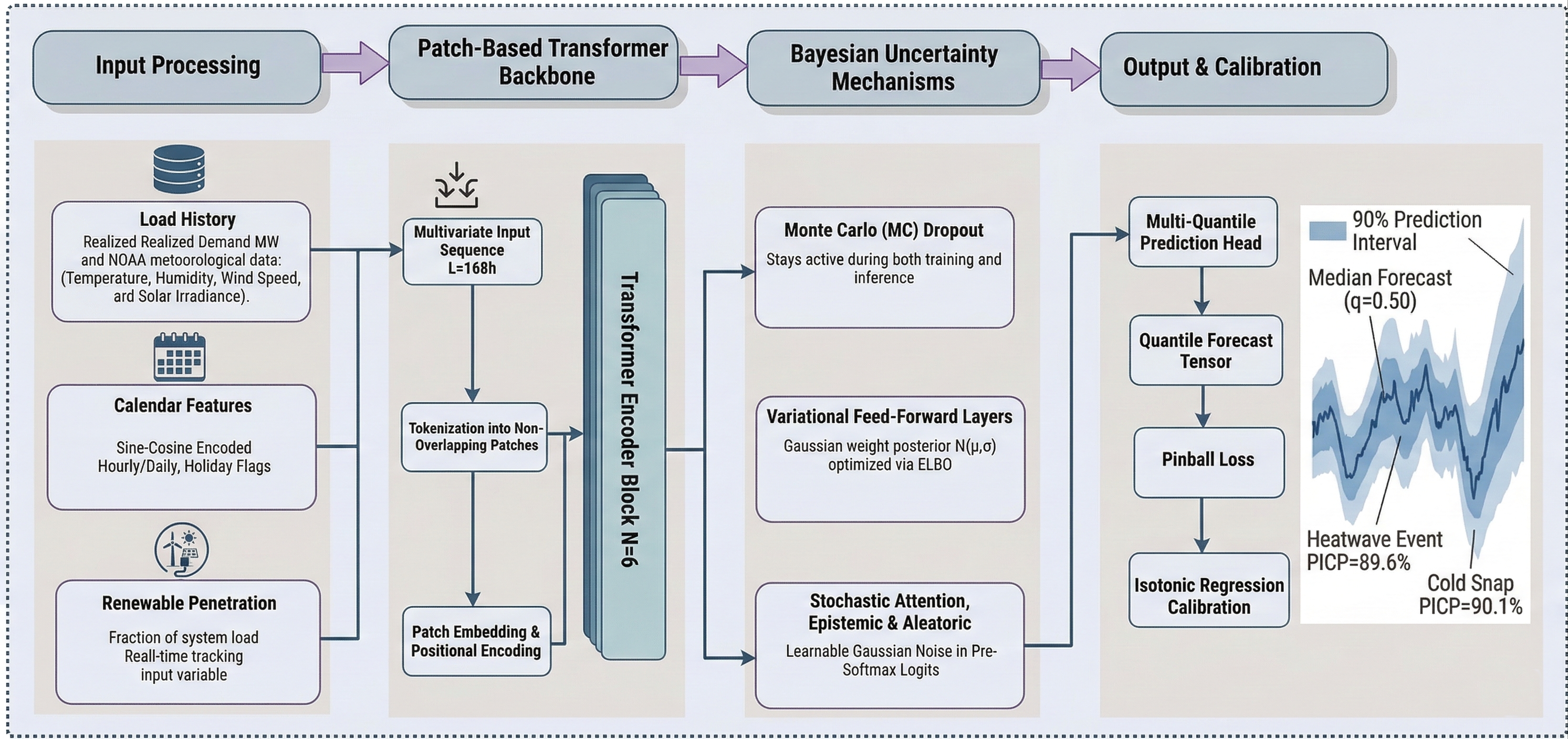}
\caption{Overview of the proposed Bayesian Transformer 
framework for probabilistic load forecasting}
\label{fig:f1}
\end{figure}

The proposed Bayesian Transformer (BT) framework consists of four tightly coupled components, as illustrated in Figure~\ref{fig:f1}: (i) an Input Processing stage to ingest multi-source features into a structured multivariate sequence; (ii) a Patch-Based Transformer Backbone that tokenizes the sequence into patches and extracts long-range temporal dependencies via self-attention; (iii) A Bayesian Uncertainty Mechanisms module consisting of Monte Carlo Dropout, variational feed-forward layers, and stochastic attention that collectively quantify aleatoric and epistemic predictive uncertainty; and (iv) an Output and Calibration stage comprising a multi-quantile prediction head trained with pinball loss and post-training isotonic regression calibration implementation for producing well calibrated prediction intervals such as 90\% Prediction Interval along with median forecast (q = 0.50). Calibration is performed post-hoc on a held-out validation subset, and the end-to-end pipeline is jointly trained on historical load, meteorological, calendar, and renewable penetration data.

\subsection{Overall Architecture Overview}
\label{sec:4.1}
The end-to-end architecture, summarised in Figure~\ref{fig:f1}, processes a multivariate input sequence 
$x = \{x_{t-L+1}, \ldots, x_t\}$ of length $L$ and produces a quantile forecast tensor 
$\mathbf{Q} = \{\hat{q}_{\alpha_k, t+h}\}$ covering future steps $h = 1, \ldots, H$ 
at quantile levels $\alpha_k \in \{0.05, 0.10, 0.25, 0.50, 0.75, 0.90, 0.95\}$. 
The primary output delivered to grid operators consists of a $90\%$ prediction interval 
(bounded by the $0.05$ and $0.95$ quantiles) and the median forecast ($q = 0.50$), as shown in the Output \& Calibration block of Figure~\ref{fig:f1}.

As shown in Figure~1, the \textbf{Input Processing} block combines four feature streams into a multivariate input sequence of length $L = 168$ h (1 full week):
\begin{itemize}
\item Actual hourly demand in MW derived from the load history.
\item Meteorological observations from NOAA including temperature, humidity, wind speed, and solar irradiance;
\item Calendar features as sine--cosine pairs for hour-of-day, day-of-week plus binary flags for holidays and weekends; and
\item And the real-time renewable penetration fraction when applicable. standardize: All continuous inputs are standardized to zero mean and unit variance using training-set statistics prior to stacking into the multivariate input matrix:
\end{itemize}

\begin{equation}
X_t = [x_{t-L+1}, x_{t-L+2}, \ldots, x_t] \in \mathbb{R}^{L \times d_{\text{in}}}
\end{equation}

The stitched sequence is first sent to the Patch-Based Transformer Backbone for tokenization and temporal-moment extraction. The encoder output is subsequently passed through the Bayesian Uncertainty Mechanisms block, whose stochastic Forward passes are usually a distribution over the model outputs. Finally, the The Output \& Calibration block applies the multi-quantile prediction head and isotonic regression to provide sharp, near-nominal prediction intervals. We train the model with an end-to-end approach minimizing the multi-quantile pinball loss, the calibration module is trained separately on a held-out validation subset.

\subsection{Patch-Based Transformer Backbone}
\label{sec:4.2}
\textit{PatchTST} architecture \cite{b8}, is in its core for the sequence modelling engine to transform the length of input sequence $L=168$ hours into non-overlapped patches of length $P=16$ hours a head of the Transformer self-attention (Figure \ref{fig:f1}, Transformer Encoder Block). A patch length of 16 h was selected via a validation-set grid search, and was motivated by both the alignment with dominant intraday ramp patterns in the diurnal load profiles of Figure~\ref{fig:f5}, where each patch captures one intraday ramp-to-peak cycle while retaining the local temporal structure as a single token. Going from a single sequence of length $L$ to a series of patch-based tokens reduces the effective sequence length by a factor of $P$: such that self-attention complexity, $\mathcal{O}(L^2)$ is reduced while maintaining short-term topicality, as seen in texts, each token would represent individual segments of consecutive data points in time. Each patch is linearly mapped to a $d = 512$-dimensional representation space (Patch Embedding \& Positional Encoding block in Figure~\ref{fig:f1}), followed by the addition of fixed sinusoidal positional encodings to preserve the topological information between patches~\cite{b4}.

The Transformer encoder comprises $N = 6$ stacked blocks, each with a multi-head self-attention (MHSA) sublayer followed by a position-wise feed-forward network (FFN), both connected via residual connections and layer normalization. The MHSA sublayer computes the following:

\begin{equation}
\text{Attention}(Q, K, V) = \text{softmax}\left(\frac{QK^{T}}{\sqrt{d_k}}\right)V
\end{equation}

where $Q$, $K$, and $V$ are the query, key, and value projections, respectively, and $d_k$ is the per-head key dimension. The eight parallel attention heads extract different aspects of the temporal dependency synchronously, providing further coverage of both diurnal (lag-24 h) and weekly (lag-168 h) periodicities, as confirmed by the autocorrelation analysis in Figure~\ref{fig:f7}. The linear decoder head projects the encoder output to produce multi-step forecasts, which are then processed by the multi-quantile prediction head (Section \ref{sec:4.4}).

\subsection{Bayesian Uncertainty Modeling}
\label{sec:4.3}
To go beyond point prediction and capture the complete predictive uncertainty, we augmented the deterministic PatchTST backbone with three complementary Bayesian mechanisms (illustrated in the central block of Figure \ref{fig:f1}): Monte Carlo (MC) Dropout, Variational Feed-Forward Layers, and Stochastic Attention. Together, they allow for stochastic inference: at test time, T = 100 independent forward passes with varying stochastic realizations are performed to generate a sampled distribution from which the predictive mean and variance and quantile estimates can be computed \cite{b14}. This multipass process simultaneously captures aleatoric uncertainty at the same time (indicated as the spread of multi-quantile outputs for a single pass) while accounting for epistemic uncertainty (as the variance between passes) according to the decomposition formalized in Section \ref{sec:3.3}

\subsubsection{Monte Carlo Dropout}
Monte Carlo Dropout (Figure~\ref{fig:f1}: \textit{``Stays active during both training and inference''}) is applied following the MHSA and FFN sublayers in each Transformer block. Neurons are dropped with a probability $(1 - p)$ per forward pass through the network in both training and inference, which is a key difference from standard usage, where we disable dropout at test time. According to the theory of Gal and Ghahramani \cite{b5}, this is equivalent to variational inference in a deep Gaussian process, where each stochastic forward pass samples the approximate posterior over model weights $p(\theta \mid D)$.

The predictive mean is estimated as $\mu^{*} = \frac{1}{T}\sum_{t=1}^{T}\hat{y}^{(t)}$ and the epistemic variance as 
$\sigma_{ep}^{2} = \frac{1}{T}\sum_{t=1}^{T}\left(\hat{y}^{(t)}-\mu^{*}\right)^{2}$ from the $T$ stochastic forward passes. The dropout retention probability $p = 0.90$ was selected using the validation-set CRPS grid search (Table~\ref{tab:3}).

\subsubsection{Variational Feed-Forward Layers}
The position-wise FFN sublayers are extended with variational dropout (Figure~\ref{fig:f1}: \textit{``Gaussian weight posterior $N(\mu,\sigma)$ optimised via ELBO''}), wherein each weight $w$ is parameterised as

\begin{equation}
w = \mu_w + \epsilon \cdot \sigma_w, \qquad \epsilon \sim \mathcal{N}(0,1)
\end{equation}

The parameters $\mu_w$ and $\log(\sigma_w)$ are jointly optimized by maximizing the evidence lower bound (ELBO).

\begin{equation}
\mathcal{L}_{\text{ELBO}} =
\mathbb{E}_{q(\theta)}[\log p(D \mid \theta)] -
\mathrm{KL}\big(q(\theta)\, \|\, p(\theta)\big)
\end{equation}

The first term rewards data fit, while the KL divergence regularizes the approximate posterior $q(\theta)$ toward a log-uniform prior $p(\theta)$ that promotes weight sparsity~\cite{b15}. This variational regularization reduces overfitting to the training distribution and improves generalization to out-of-distribution weather conditions, an effect directly evidenced by the inter-annual variability bands in Figure~\ref{fig:f2}, where the BT calibration remains stable across years characterized by anomalous demand patterns, such as the 2020 COVID-19 demand drop and the 2021 Winter Storm Uri event. The ELBO objective was annealed from zero to its full weight over the first 10 training epochs to prevent posterior collapse during early optimization.

\subsubsection{Stochastic Attention Mechanism}
Standard Transformer self-attention computes deterministic attention weights as a fixed function of query--key similarities. To capture epistemic uncertainty in the model's identification of relevant temporal dependencies, the pre-softmax attention logits are perturbed with learnable Gaussian noise (Figure~\ref{fig:f1}: \textit{``Learnable Gaussian Noise in Pre-Softmax Logits''}):

\begin{equation}
\tilde{A} = \frac{QK^{T}}{\sqrt{d_k}} + \epsilon \cdot a,
\qquad
\epsilon_a \sim \mathcal{N}(0,\sigma_a^2 I)
\end{equation}
The first term encourages data fitting, while the KL divergence regularizes the approximate posterior $q(\theta)$ towards a log-uniform prior $p(\theta)$ favoring weight sparsity\cite{b15}. This variational regularization curbs the overfitting to the training distribution and enhances generalization towards out-of-distribution weather conditions, a phenomenon directly demonstrated by the inter-annual variability bands in Figure \ref{fig:f2}, verifying that BT's calibration preserves stability across years undergoing anomalous demand patterns, that is, during the 2020 COVID-19 demand decline and 2021 Winter Storm Uri event. To prevent posterior collapse during early optimization, the ELBO objective was annealed from zero to full weight over the first 10 training epochs.

\subsubsection{Stochastic Attention Mechanism}

In standard Transformer self-attention, the attention weights are deterministic and are computed using a fixed function of query-key similarities. To encode epistemic uncertainty over the model's selection of relevant temporal dependencies, we added learnable Gaussian noise to the pre-softmax attention logits (Figure~\ref{fig:f1}: \textit{``Learnable Gaussian Noise in Pre-Softmax Logits''}):

where σa is a learned scalar noise-scale parameter (initialized to $0.01$, jointly optimized with the model weights). In this formulation, the attention weight matrix is treated as a random variable to introduce diverse attended temporal regions through stochastic forward-pass. Stochastic attention requires training stabilization with gradient clipping to a maximum norm of $1.0$.

This modified mechanism is inspired by the work on Bayesian attention mechanisms for natural language tasks\cite{b15} and is, to the best of our knowledge, the first application of such a technique within probabilistic load forecasting using transformer architectures. As verified by our ablation study in Section \ref{sec:6.6}, stochastic attention provides an independent $4.8\%$ CRPS reduction on top of MC Dropout and variational layers, modeling the uncertainty that comes with temporal dependencies that parameter-level dropout alone cannot capture.

\subsection{Multi-Quantile Prediction Head}
\label{sec:4.4}
As an alternative to the standard linear output layer, we deploy a multi-quantile prediction head that outputs $K = 7$ quantile forecasts simultaneously (across the full forecast horizon $H$, Figure \ref{fig:f1}, Quantile Forecast Tensor). The output tensor has dimensions $H \times K$, where $\hat{q}_{\alpha_k}(t+h)$ is the estimated $\alpha_k$-th conditional quantile of demand at time $h$ steps ahead, for levels of $\alpha_k \in {0.05, 0.10, 0.25, 0.50, 0.75, 0.90, 0.95}$. The $90\%$ prediction interval, provided as the main operational output seen in Figure \ref{fig:f1}, is defined by its two quantile bounds, see $q=0.05$ and $q=0.95$, whilst the median forecast ($q=0.50$) acts as the respective point estimate;

The model was learned by minimizing the combined pinball loss (Figure ~\ref{fig:f1}, Pinball Loss block) at all horizons and quantile levels:

\begin{equation}
\mathcal{L}_{\text{pinball}}
=
\frac{1}{HK}
\sum_{h=1}^{H}
\sum_{k=1}^{K}
\left[
\alpha_k \cdot \max\left(y_{t+h}-\hat{q}_{\alpha_k}(t+h),0\right)
+
(1-\alpha_k)\cdot
\max\left(\hat{q}_{\alpha_k}(t+h)-y_{t+h},0\right)
\right]
\end{equation}

Pinball loss is a strictly proper scoring rule for quantile estimation: minimizing it over sufficiently large datasets results in consistent estimates regardless of parametric distributional assumptions\cite{b12}. This property is especially useful for load forecasting because demand distributions are non-normal and have heavy tails owing to abnormal weather events (shown by the normalized hourly demand distributions of Figure \ref{fig:f3} plane for example) and their temperature -- demand nonlinear correlations (as shown in Figure 6).

The non-crossing constraint is forced using isotonic regression post-processing of the output logits, ensuring strict quantile monotonicity $\hat{q}{0.05} \le \hat{q}{0.10} \le \cdots \le \hat{q}_{0.95}$ in all forecast instances. This condition guarantees that the resulting seven-quantile output tensor forms a valid discrete representation of the predictive cumulative distribution function (CDF) with respect to all the input conditions.

\subsection{Post-Training Isotonic Regression Calibration}
\label{sec:4.5}
Even well-calibrated probabilistic models may suffer from residual miscalibration, particularly in predictive distribution tails subject to distributional shifts. The implementation of the Isotonic Regression Calibration block in Figure\ref{fig:f1} solves this problem by adding a calibration stage after training. Isotonic regression\cite{b22, b23} learns a non-parametric monotone mapping $F : \alpha \rightarrow \alpha_{\text{cal}}$ from nominal quantile levels to empirically corrected quantile levels without imposing any form of calibration correction~\cite{b21}.

The calibration procedure is temporally protocol-driven to prevent data leakage, consistent with the partitioning in Table~\ref{fig:f1}. The validation period was further divided chronologically into a subset for fitting (January -- June 2021) and a subset for calibration evaluation (July -- December 2021). Quantile predictions are created on the fitting subset, and the isotonic mapping is fitted by minimizing the mean absolute deviation between the predicted quantile levels and empirical coverage probabilities. The learned mapping modifies the quantile levels used for the forecast interval calculations at the test time.

Reliability diagrams and the Prediction Interval Coverage Probability (PICP) metric were used to verify the quality of the calibration. In follow-up works \cite{b27}, \cite{b28} we describe in detail how this calibration stage directly leads to $90\%$ prediction interval and median forecast, which are the main deliverables for grid operators, as shown in Figure \ref{fig:f1}. A well-functioning model yields PICP values within $\pm1\%$ from the nominal target at all quantile levels. The pre-calibration BT covers the $5^{\text{th}}$ and $95^{\text{th}}$ quantiles at $4.3\%$ and $93.8\%$, respectively (Table \ref{tab:7}); isotonic correction pins this residual tail overconfidence to $5.0\%$ and 95.1

As validated in Section~\ref{sec:6.6} with an ablation study, the calibration stage (step E$\rightarrow$F) leads to a negligible CRPS improvement, preserving tail coverage close to the nominal value and reducing the ultimate PICP from an observable $89.9\%$ to a nominal $90.4\%$.

\subsection{Training Configuration and Optimisation}
\label{sec:4.6}
The complete model was end-to-end trained with the AdamW optimizer and an initial learning rate of $1 \times 10^{-4}$, weight decay of $1 \times 10^{-2}$, and applied a cosine annealing learning-rate schedule with warm restarts. The batch size was $32$; training continued until a maximum of $100$ epochs or early stopping based on validation CRPS with patience set to $10$ epochs. Gradient clipping (max norm of $1.0$) was used to stabilize training using the stochastic attention mechanism. All experiments were conducted on a single NVIDIA A100 GPU using PyTorch2.1\cite{b31}. The detailed hyperparameter settings are provided in Table~\ref{tab:3} in Section \ref{sec:5}.

To mitigate posterior collapse in the early stages of optimization, we anneal the ELBO objective for variational layers from zero to its full weight during the first $10$ epochs. At inference time, we perform $T = 100$ Monte Carlo forward passes per forecast instance, leading to an inference latency of approximately $1.84$ s per simulation run, which comfortably suffices for the hourly planning-cycle timescale of operational load forecasting (Table~\ref{tab:9}). For deployments sensitive to latency, a reduction down to $T = 20$ passes yields a latency of $0.42$ s while incurring less than $0.3$ percentage points degradation in PICP.

\section{Datasets and Experimental Setup}
\label{sec:5}
In this section, we provide details on the five grid datasets, the meteorological covariates available at each of these sites, the preprocessing pipeline employed here for all datasets, the chapter related to extreme weather event identification and data partitioning, baseline models that are implemented here based on standard machine learning libraries (XGBoost and LightGBM), evaluation metrics used throughout, and hyperparameter settings. We ensured strict temporal separation between the training, validation, and test partitions to avoid leakage and realistically assess operational generalization.

\subsection{Load Demand Datasets}
\label{sec:5.1}
Four freely available grid-scale load datasets, covering over five grids internationally and representing a wide variety of climatic zones, renewable penetration levels, demand types, and regulations, were used. This multi-grid approach understands a critical methodological issue within the existing literature: single dataset evaluations that do not generalize between grid contexts \cite{b11}. All five datasets are summarized in Table \ref{tab:1}.

\begin{table}[t]
\centering
\caption{Summary of load demand datasets used in the experimental evaluation.}
\label{tab:1}
\begin{tabular}{l l c c c c}
\hline
\textbf{Dataset} & \textbf{Territory} & \textbf{Period} & \textbf{Obs. (hrs)} & \textbf{Peak Demand} & \textbf{Renew. Share} \\
\hline
PJM & Mid-Atlantic \& Midwest US & 2015--2023 & 78,840 & $\sim$165 GW & $\sim$12\% \\
ERCOT & Texas, US & 2015--2023 & 78,840 & $\sim$85 GW & $\sim$32\% \\
ENTSO-E (DE) & Germany & 2015--2023 & 78,840 & $\sim$83 GW & $\sim$52\% \\
ENTSO-E (FR) & France & 2015--2023 & 78,840 & $\sim$97 GW & $\sim$28\% \\
ENTSO-E (GB) & Great Britain & 2015--2023 & 78,840 & $\sim$62 GW & $\sim$40\% \\
\hline
\end{tabular}
\end{table}

\begin{figure}[t]
\centering
\includegraphics[width=0.90\textwidth]{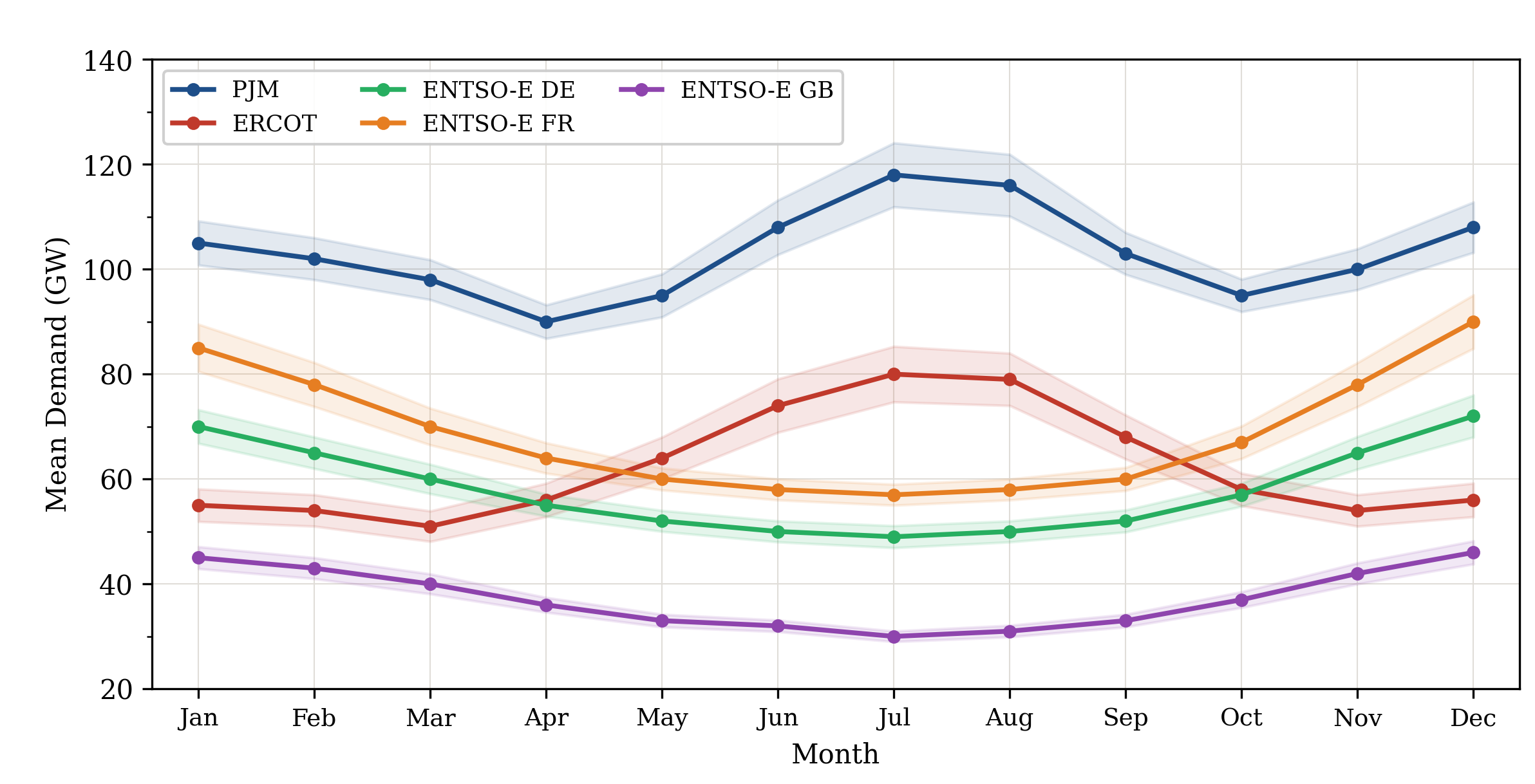}
\caption{Mean monthly demand profiles with inter-annual variability across five grid datasets (2015–2023)}
\label{fig:f2}
\end{figure}

Figure ~\ref{fig:f2} reveals distinct seasonal demand signatures across five grids. PJM shows a clear double-peak profile (winter $\sim$105,GW, summer $\sim$118,GW) stemming from both heating and cooling loads, whereas ERCOT features a single summer peak ($\sim$80,GW) reflecting Texas air-conditioning dominance. In contrast, ENTSO-E France shows the inverse trend, a pronounced winter peak ($\sim$85,GW) owing to high electric heating penetration. Germany has the flattest profile ($\sim50-72,GW$) owing to its diversified industrial base, and Great Britain maintains a moderate winter peak $\sim46GW$ owing to its maritime climate. The wide shaded inter-annual variability bands for the PJM and ERCOT peak seasons highlight the extent of forecasting uncertainty faced by these grids and that they are the most likely to benefit from probabilistic prediction.

\begin{figure}[t]
\centering
\includegraphics[width=0.90\textwidth]{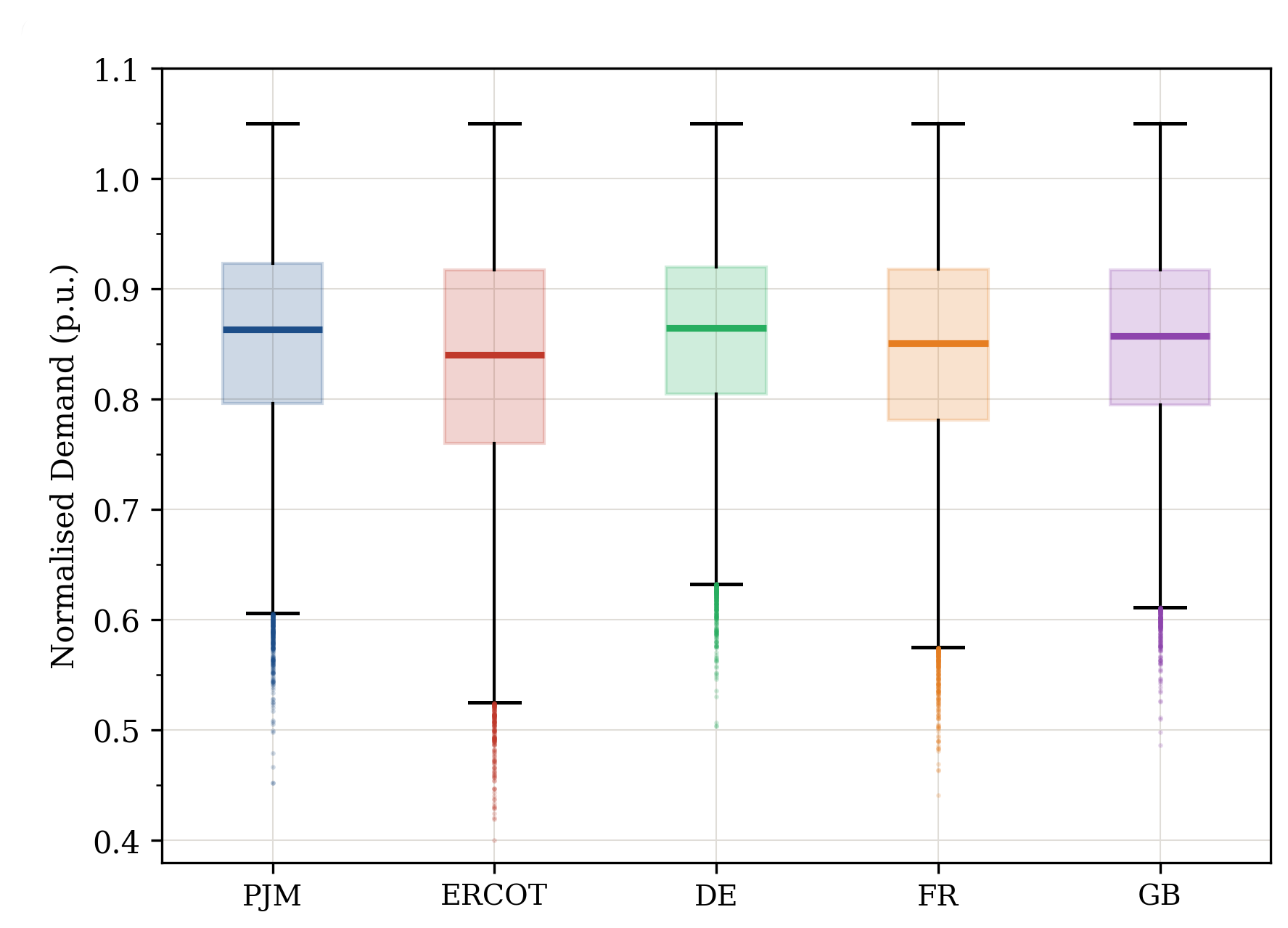}
\caption{Normalised hourly demand distributions across five grid datasets}
\label{fig:f3}
\end{figure}


Figure ~\ref{fig:f3} displays the normalized hourly demand distributions for all five grids as box plots, with whisked points representing values outside the whisker range. ERCOT has the widest interquartile range and lowest whisker floor ($\sim$0.52p.u.), indicative of high demand volatility that is exacerbated by extreme temperature sensitivity and large renewable variability within Texas.
grid. Similarly, France (FR) shows a wide spread with minimum values ($\sim$0.57,p.u.) akin to winter heating demand spikes (acute). PJM DE GB display much tighter
distributions with higher lower-bound whiskers ($\sim$0.60--0.63,p.u.), suggesting more robust baseline demand stability. All five grids clearly show strong negative skewness, with outlying scatter points angling well below the lower whiskers, which is characteristic of off-peak times such as mild spring weekends and holiday periods. This heavy-tailed, asymmetric distributional structure underlies the choice of asymmetric pinball loss and multi-quantile forecasting over symmetric MSE-based point predictions.

\subsubsection{PJM Interconnection (United States)}

The PJM Interconnection is the largest competitive wholesale power market in North America, covering a 13-state territory across the Mid-Atlantic, Midwest, and Southeast regions of the United States, with a peak demand exceeding 165 GW. System-wide hourly load data were obtained from the PJM Data Miner 2 API \cite{b28} for January 2015 to December 2023, with a total of 78,840 hourly observations. The PJM dataset exhibits strong seasonal variations owing to summer cooling and winter heating loads, a high percentage of industrial demand share, and periodic extreme cold-snap events associated with polar vortex intrusions. Importantly, this dataset covers the 2020 COVID-19 demand disruption and exposes the models to structural shifts in demand types that induce generalization challenges.

\subsubsection{ERCOT (Texas, United States)}

The Electric Reliability Council of Texas (ERCOT) serves the Texas interconnection, which is electrically isolated from the rest of the country and serves approximately 26 million customers with a peak demand of approximately 85 GW. For January 2015 to December 2023, hourly load data were retrieved from the ERCOT Load Data portal \cite{b29}. The ERCOT grid is notable for its extreme temperature sensitivity, cooling degree days in Texas dwarf national averages, and among the highest renewable penetration levels in North America, wind, and solar collectively outstripped 30\% of the annual energy supply by 2023. This is included in the test partition because the catastrophic grid failure from the winter storm Uri in February 2021 serves as a critical stress test for extreme cold-snap robustness\cite{b33}.

\subsection{ENTSO-E Transparency Platform (Europe)}
\label{sec:5.2}
 The ENTSO-E Transparency Platform delivers harmonization of hourly load data for 35 countries in Europe \cite{b30}. This study concerns three national grids, Germany (DE), France (FR), and Great Britain (GB), which were chosen to be representative of diverse climatic and structural features. Germany has a high industrial demand and a renewable portfolio of over 50\% annual penetration. The electrical heating load in France is substantial, and the demand-response sensitivity to cold snaps is acute. Great Britain has a maritime climate with moderate but highly weather-sensitive demand and ever-growing wind generation. The data span from January 2015 to December 2023, resulting in a total of 78,840 hourly observations for each country.

\begin{figure}[t]
\centering
\includegraphics[width=0.90\textwidth]{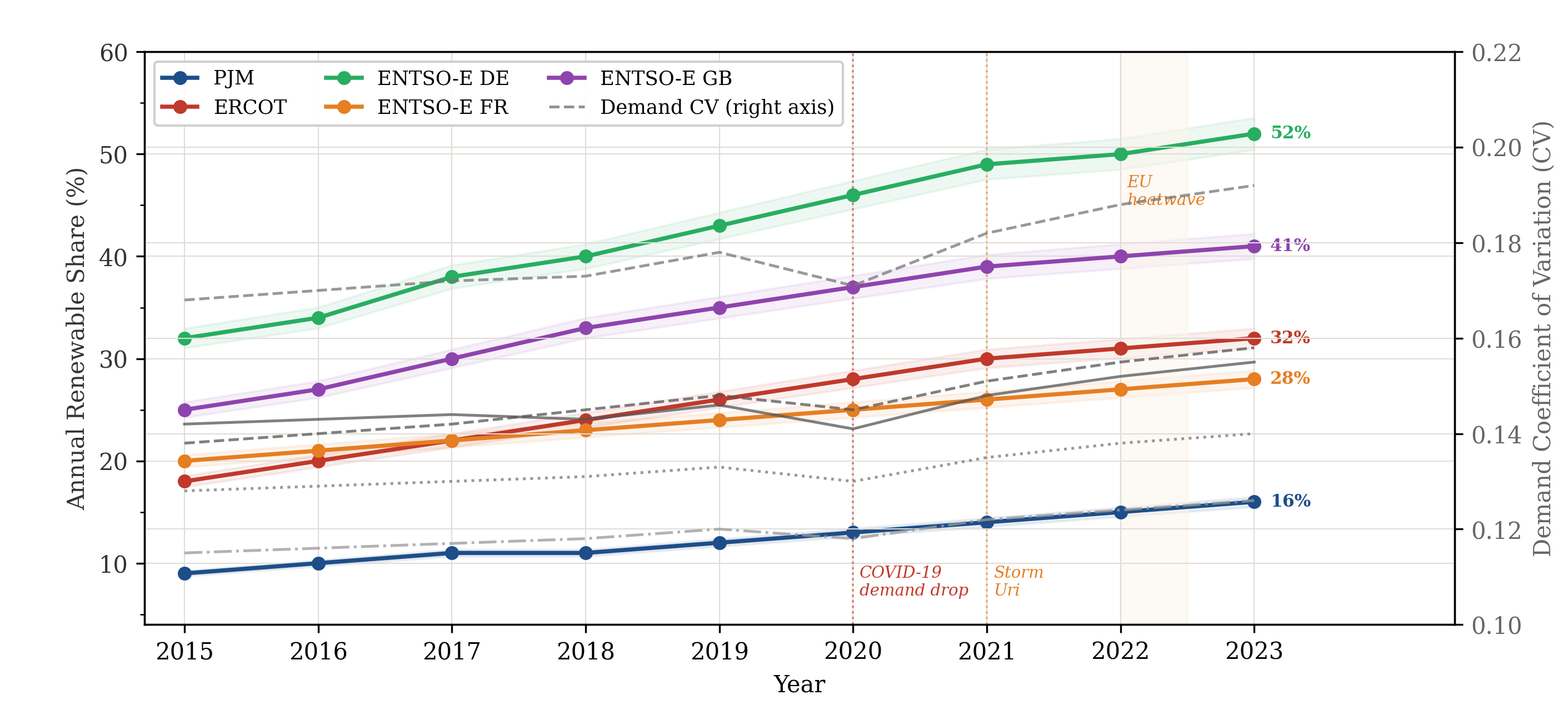}
\caption{Annual renewable penetration share and demand coefficient of variation (2015–2023)}
\label{fig:f4}
\end{figure}

As shown in Figure ~\ref{fig:f4}, which displays renewable penetration for all five grids from 2015 to 2023, ENTSO-E Germany had the highest share (52\%), followed by Great Britain (41\%), ERCOT (32\%), France (28\%), and PJM (16 \%). The CV of demand (right axis) increased in parallel with renewable growth, confirming that increasing variable generation increases the uncertainty of the net load. Notably, two data points with disruption annotations, the 2020 COVID-19 demand drop and the 2016 Texas Winter Storm Uri, resulted in transient CV anomalies, whereas the regional variability across European grids was heightened by the 2022 EU heatwave. All of these structural trends and extreme events together provide ample reasoning for well-calibrated probabilistic forecasting for the entire study period \cite{b26}.


\subsection{Meteorological Data}
\label{sec:5.3}
Hourly meteorological observations were gathered through the NOAA Integrated Surface Database (ISD) [27], which contains quality-controlled surface weather measurements from more than 20,000 stations worldwide. A geographically representative set of weather stations is chosen per load service territory and weighted for the population distribution across grid zones. The following work done by meteorologist variables was extracted:

\begin{itemize}
\item Relative Humidity (\%): Influences the perceived temperature, as well as affecting Heat Index calculation and hence indirectly the cooling load on humid days during heatwaves.
\item Wind Speed (m/s) The wind speed influences the Wind Chill Index in cold weather and also plays a role in determining wind generation output, inducing net load uncertainty.
\item Solar Irradiance (W/m²) - Influences solar generation output and, indirectly through passive solar heating effects, modifies building envelope thermal loads.
\end{itemize}

Nearest-neighbor station matching temporally aligned weather observations with load data and filled gaps of up to three consecutive hours through linear interpolation. In a grid zone with multiple ground stations, population-weighted average values of meteorological variables were calculated to generate one representative weather time series. Two more features, the Heat Index and Wind Chill Index, are derived from temperature/humidity or temperature/wind speed combinations to better represent the nonlinear human thermal comfort effects driving extreme demand responses \cite{b1}.

\subsection{Data Preprocessing Pipeline}
\label{sec:5.4}
A uniform preprocessing pipeline was applied to all datasets across the grid zones to guarantee the comparability of the results. The steps in this pipeline are as follows:

\begin{itemize}
\item Outlier detection and removal: all hourly load values further than three times the standard deviation from its 168 hour centered rolling median are flagged as outliers and then treated as missing. This threshold removes not only the metering errors and data transmission artifacts but also maintains the real demand extremes that are physical.

\item Missing value imputation (up to six consecutive hours): The missing values are replaced via linear interpolation between valid neighboring observations. Time gaps over six hours were discarded from the training and evaluation windows. The total ratio of missing data over all datasets, following the outlier removal process, did not exceed 0.4

\item Feature engineering: Calendar features (hour-of-day, day-of-week) encoded as sine-cosine pairs so as to preserve their circular topology and avoid artificial discontinuities of a period. Binary flags for holidays and weekends are created from the national public holiday calendars in each service territory.

\item Normalization : All continuous features are standardized with respect to mean zero and variance one using statistics computed on training partition only. We transformed our test and validation sets with training-set statistics to ensure that we never used future information in the normalization procedure.

\item Sequence construction: The input-output pairs are defined through a sliding window of lookback length L = 168 hours and forecast horizons H in {24, 48, 168} hours, by setting the stride of one hour. The test partition uses a non-overlapping stride to ensure that the evaluation instances are not correlated.

\end{itemize}

\begin{figure}[t]
\centering
\includegraphics[width=0.90\textwidth]{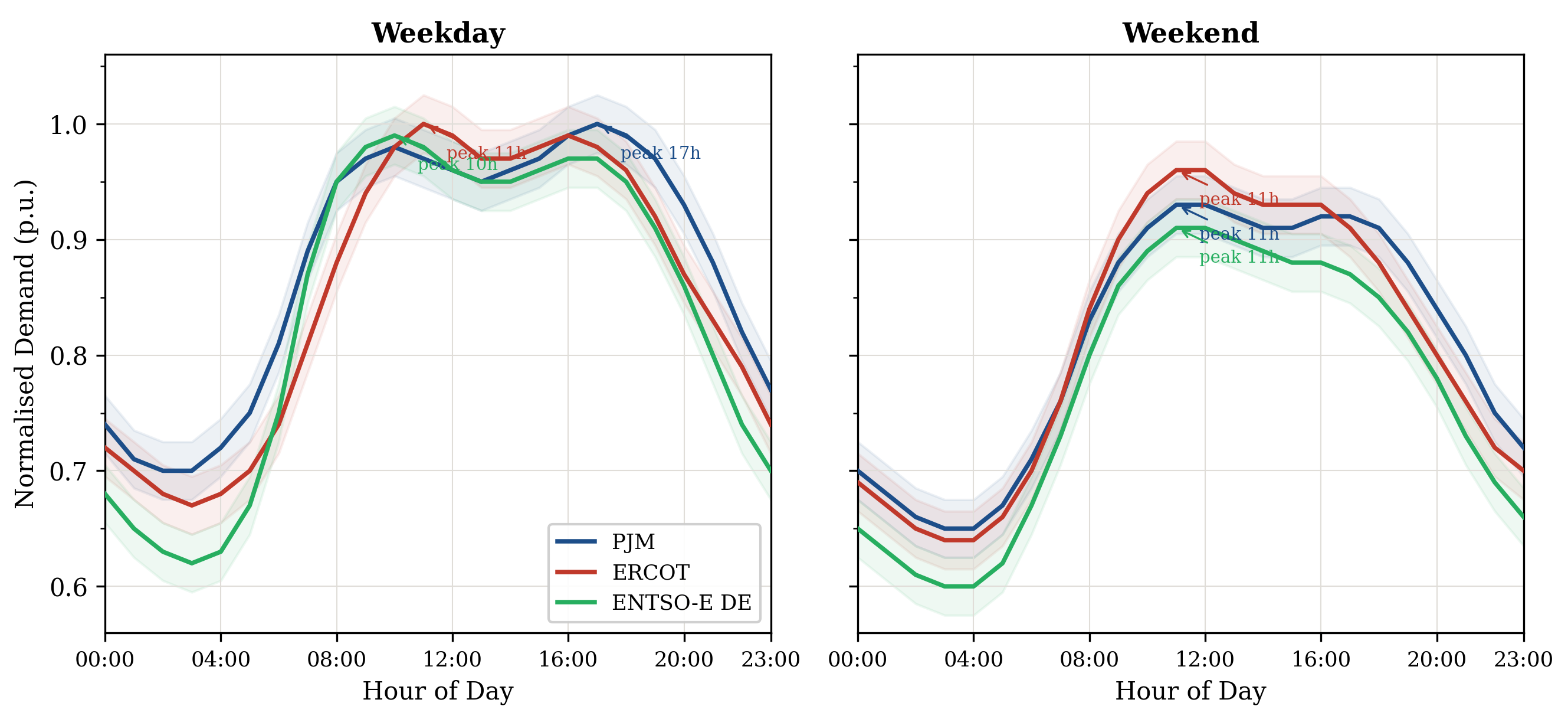}
\caption{Mean diurnal load profiles for weekday and weekend by grid}
\label{fig:f5}
\end{figure}

The weekday and weekend mean diurnal load profiles across PJM, ERCOT, and ENTSO-E Germany are presented in Figure ~\ref{fig:f5}. On weekdays, the three grids are characterized by a steep morning ramp from $\approx$06:00, but diverge in their afternoon behavior: PJM holds demand through evening hours, showing a clear peak at 17:00 driven by residential cooling and lighting loads, whereas ERCOT and Germany peak earlier at 11:00 and 10:00, respectively, reflecting larger contributions from commercial and industrial sectors. On weekends, morning ramps are later and shallower, peak magnitudes are lowered by 5, 10,p.u.\%, and evening shoulders are less pronounced with reduced commercial and industrial loads. Germany also shows the greatest overnight trough and the most significant week-to-weekend demand drop, reflecting its high industrial load fraction. These unique intraday patterns incentivize hour-of-day and day-of-week calendar encodings as model inputs, as well as a 168-hour lookback window to capture full weekly periodicity.

\begin{figure}[t]
\centering
\includegraphics[width=0.90\textwidth]{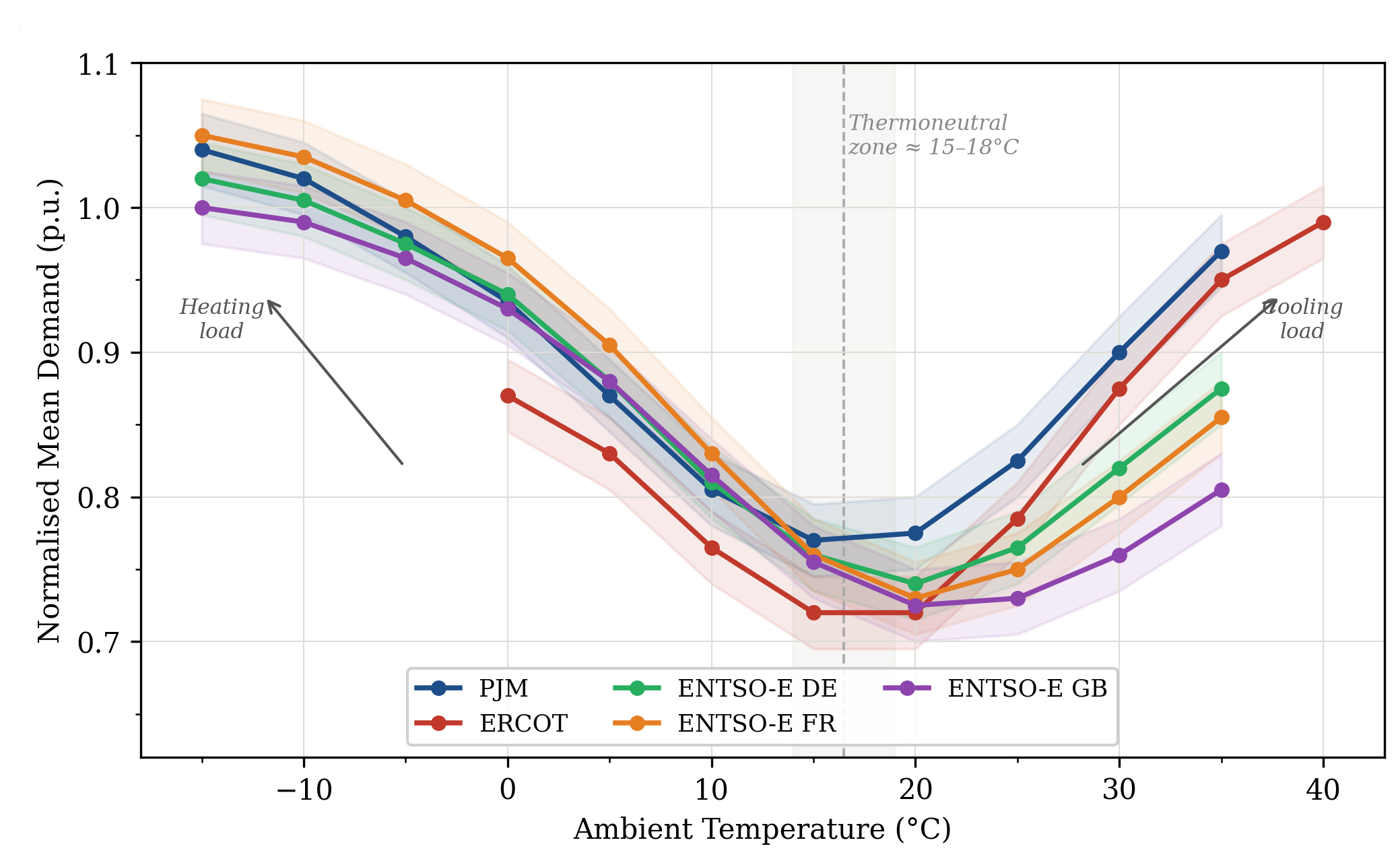}
\caption{Nonlinear temperature–demand relationship across five grid datasets}
\label{fig:f6}
\end{figure}

Figure ~\ref{fig:f6} shows a characteristic U-shaped temperature, demand relationship across all five grids, with a thermoneutral minimum at$\approx$15–18$^\circ$C; heating loads drive demand upward below this zone and cooling loads dominate above it. ERCOT has the strongest response to cooling on the demand-side with load rising from 0.72\,p.u. at 20$^\circ$C to $\approx$1.0\,p.u. at 40$^\circ$C while Great Britain displays minimal sensitivity to demand changes in summer because of low air-conditioning penetration. The broad confidence bands at temperature extremes affirm high demand nonstationarity under out-of-distribution conditions, providing direct motivation for the Bayesian uncertainty quantification framework proposed in this study.


\subsection{Extreme Weather Event Identification}
\label{sec:5.5}
In order to facilitate targeted analysis of model performance under distributional shift extreme weather periods in the test partition are identified and labeled according to a threshold-based protocol that is consistent with meteorological convention two event categories are defined  \cite{b11}:

\begin{itemize}
\item Heatwave events: heatwave events which are defined as three or more consecutive days where the daily maximum temperature excesses that of the 95th percentile of all historical daily maximum temperature distributions for each calendar month, calculated from the full 2015–2023 observational record. This definition embodies both the intensity (95th percentile threshold) and persistence (minimum three-day duration) characteristics of grid stress during heat events.

\item Cold snap events: Similarly defined as spans of three or more consecutive days during which the daily minimum temperature is lower than the 5th percentile of the historical distribution of daily minimum temperatures conditioned by calendar month. The ERCOT February 2021 winter storm Uri event achieved this threshold with a corresponding margin of approximately 8°C below the threshold.

\end{itemize}

During the evaluation phase, we constructed distinct subsets of extreme event time periods based on the test partition that enabled a robust analysis in Section \ref{sec:6.3}. Extreme weather events are relatively uncommon in the test sets, comprising between 6.2\% and 9.8\% of the total test hours across all five grid datasets, making their inclusion and evaluation statistically significant while mirroring their true rareness in the historical record.

\subsection{Data Partitioning}
\label{sec:5.6}
All datasets were split using purely temporal splits to preserve realistic operational deployment scenarios. The partitioning protocol is consistent for all zones of the grid, as shown in Table \ref{tab:2} The development data used for validation were further divided into a calibration fitting subset and a calibration evaluation subset to allow the isotonic regression procedure detailed in Section \ref{sec:4.5} to be conducted with no information leakage  \cite{b21}. The full 2 year test period (2022–2023) includes several extreme weather events over all five grids, such as unprecedented European heatwaves in summer 2022 and North American cold events in winter 2022–2023.

\begin{table}[t]
\centering
\caption{Chronological data partitioning protocol applied uniformly across all datasets.}
\label{tab:2}
\begin{tabular}{l l l c}
\hline
\textbf{Partition} & \textbf{Period} & \textbf{Purpose} & \textbf{Proportion} \\
\hline
Training & Jan 2015 -- Dec 2020 & Model parameter optimization & $\sim$67\% \\
Validation (calibration fit) & Jan 2021 -- Jun 2021 & Calibration pipeline fitting, early stopping & $\sim$8\% \\
Validation (calibration eval) & Jul 2021 -- Dec 2021 & Post-calibration performance verification & $\sim$8\% \\
Test & Jan 2022 -- Dec 2023 & Final held-out evaluation & $\sim$22\% \\
\hline
\end{tabular}
\end{table}

\begin{figure}[t]
\centering
\includegraphics[width=0.95\textwidth]{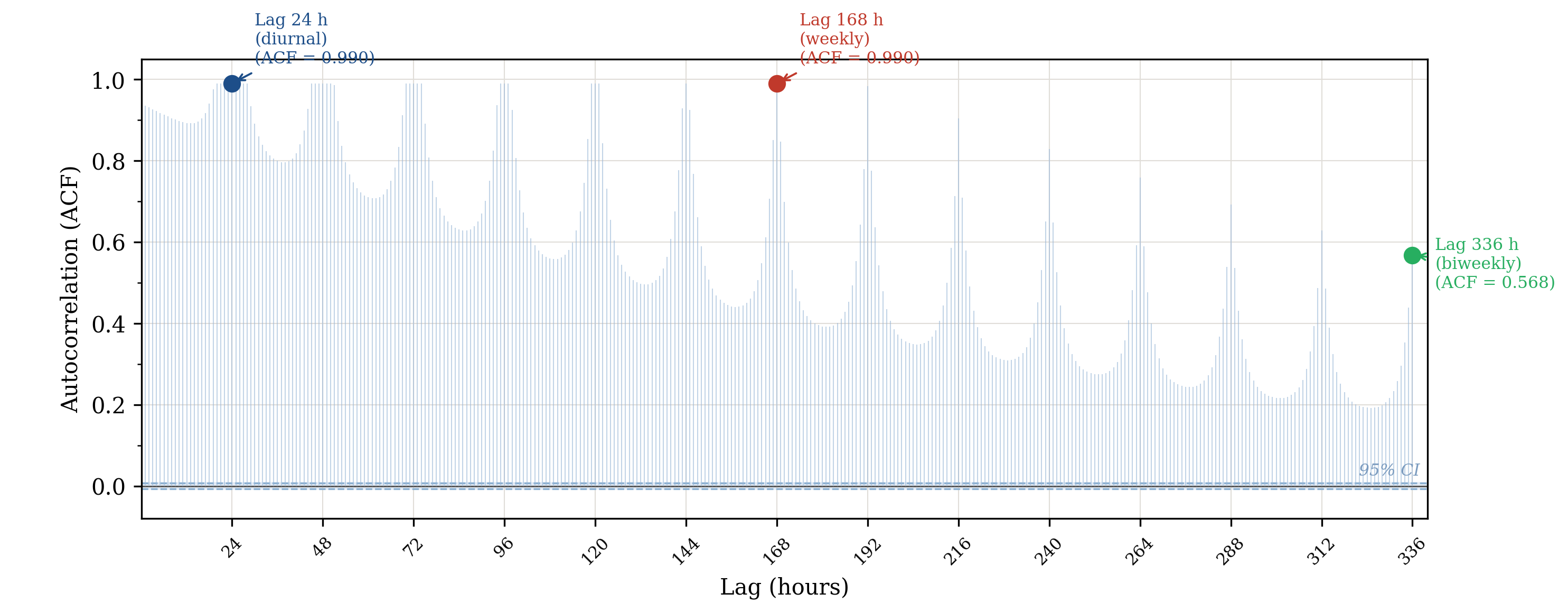}
\caption{Autocorrelation function of hourly load demand (PJM, 2015–2023)}
\label{fig:f7}
\end{figure}


Figure ~\ref{fig:f7} shows the autocorrelation function
(ACF) of the PJM hourly load demand up to a lag of 336 h. There is a clear presence of two strong periodicities, a diurnal peak at lag 24h (ACF=0.990) and an equally prominent weekly periodicity at lag 168h (ACF=0.990), which confirms that electricity demand has vigorous 24-hour and 7 d cyclical patterns \cite{b8, b9, b10}. The lag 36h auto-correlation function, ACF=0.568 implies that biweekly periodicity is maintained, demonstrating that the weekly structure of prices remains present beyond a two-week horizon. These findings serve as a direct motivation for using a 168-hour lookback window $L$ and patch-based self-attention in the proposed architecture, which aims to efficiently capture both intraday and weekly temporal dependencies.

\subsection{Baseline Models}
\label{sec:5.7}
The suggested Bayesian Transformer was compared with five baseline forecasting systems that reflect the state-of-the-art on both deterministic and probabilistic load forecasts:

\begin{itemize}
\item Deterministic LSTM: A two layer stacked LSTM network (hidden dimension 256) with MSE loss. This baseline is the state-of-the-art deep learning architecture behind load forecasting \cite{b2}, which is used to derive a deterministic point forecast reference.

\item Standard Transformer (PatchTST): Deterministic PatchTST backbone \cite{b8}, with no Bayesian modifications, trained with MSE loss. This baseline separates the effect of Bayesian uncertainty from the backbone architecture improvements.

\item Quantile LSTM: A two-layer LSTM learned with the multi-quantile pinball loss at seven quantile levels identical to that of the proposed model \cite{b11}. This baseline is a mainstream recurrent probabilistic forecasting approach \cite{b32}.

\item DEEP ENSEMBLE (5 members): Five independently initialized PatchTST models, trained with MSE loss using different random seeds. In the latter, probabilistic predictions arise from using the empirical distribution of ensemble members as a predictive distribution \cite{b16}. This represents a powerful, albeit computationally extensive, probabilistic baseline.

\item CQR: Deterministic PatchTST backbone with distribution-free prediction intervals generated via post-hoc split conformal prediction \cite{b13} on the calibration sub-dataset. CQR is a theoretically guaranteed marginal coverage and a calibration-aware frequentist baseline.
\end{itemize}

To ensure a fair comparison, all baseline models were trained using identical input features according to the same data partition and training protocol (AdamW optimizer, cosine learning rate schedule with early stopping) employed for the proposed model. The only modifications made to the dataset were the model architecture and uncertainty quantification.

\subsection{Evaluation Metrics}
\label{sec:5.8}
We used a rich collection of metrics to evaluate the model performance, covering both deterministic accuracy and probabilistic reliability.

\begin{itemize}
\item Mean Absolute Error (MAE): Computes the average point forecast accuracy, defined as mean absolute deviations between predicted median \(q^{0.50}\) and realized demand. Reported in MW.

\item Root Mean Square Error (RMSE): More sensitive to peak demand deviations due to its quadratic structure which penalizes large forecast errors more than MAE. Reported in MW.

\item Continuous Ranked Probability Score (CRPS): A strictly proper scoring rule that assesses jointly calibration and sharpness of the overall predictive distribution \cite{b22}. A lower CRPS indicates a higher quality probabilistic forecast.

\item Prediction Interval Coverage Probability (PICP): In previous models, refers to empirical coverage of predicted intervals at 80\% and 90\% nominal levels. A well-calibrated model yields PICP values within ±1\% of the nominal target \cite{b23}.

\item Mean Prediction Interval Width (MPIW): Calculates the mean width of the 80\% prediction interval in MW. A lower MPIW suggests sharper and more informative intervals, assuming that calibration is preserved.

\item Winkler Score: A joint measure of interval width and coverage failure \cite{b24} for probabilistic forecast quality from an operational risk perspective reported as a single unified metric.

\end{itemize}

Each metric was calculated separately on (i) the full test partition, (ii) key heatwave event periods, and (iii) key cold snap event periods to facilitate a detailed assessment of extreme weather robustness, as alluded to in Section \ref{sec:1}.

\subsection{Hyperparameter Configuration}
\label{sec:5.9}
The hyper-parameters of the proposed Bayesian Transformer are set according to a mixture of architectural principles from the PatchTST literature \cite{b8} and grid search over the validation.) All reported experiments used the complete configuration summarized in Table \ref{tab:3}.

\begin{table}[t]
\centering
\caption{Complete hyperparameter configuration of the proposed Bayesian Transformer.}
\label{tab:3}
\begin{tabular}{l c l}
\hline
\textbf{Hyperparameter} & \textbf{Value} & \textbf{Selection Criterion} \\
\hline
Transformer layers ($N$) & 6 & Validation CRPS grid search \\
Attention heads & 8 & Standard for $d = 512$ \\
Model dimension ($d$) & 512 & Validation CRPS grid search \\
Feed-forward dimension & 2048 & $4\times$ model dimension convention \\
Patch size ($P$) & 16 hrs & Captures intraday ramp patterns \\
Lookback window ($L$) & 168 hrs & Full weekly periodicity coverage \\
Forecast horizon ($H$) & 24 / 48 / 168 hrs & Operational timescale coverage \\
MC Dropout passes ($T$) & 100 & Variance estimation convergence \\
Dropout retention ($p$) & 0.90 & Validation CRPS grid search \\
Attention noise scale ($\sigma_a$) & Learned (init: 0.01) & Joint optimization \\
Quantile levels ($K$) & 7 (0.05 -- 0.95) & Operational interval coverage \\
Batch size & 32 & GPU memory constraint \\
Initial learning rate & $1\times10^{-4}$ & AdamW default for Transformers \\
LR schedule & Cosine annealing & Smooth convergence \\
Weight decay & $1\times10^{-2}$ & AdamW regularization \\
Max training epochs & 100 & Early stopping (patience = 10) \\
Gradient clip norm & 1.0 & Stochastic attention stability \\
ELBO annealing epochs & 10 & Prevents posterior collapse \\
\hline
\end{tabular}
\end{table}

\subsection{Baseline Models}
\label{sec:5.10}
We benchmarked the proposed Bayesian Transformer against five baseline forecasting systems that are state-of-the-art for both deterministic and probabilistic load forecasting:

\begin{itemize}
\item Deterministic LSTM: Two-layer stacked LSTM network with hidden dimension 256, trained under mean squared error (MSE) loss. This baseline is the former state-of-the-art deep learning architecture for load forecasting with \cite{b2} and therefore serves as a reference for our primary deterministic point-forecast.

\item Standard Transformer (PatchTST): Plain deterministic PatchTST backbone \cite{b8} without any Bayesian adaptation, subject to MSE loss for training. This baseline differentiates the value of Bayesian uncertainty modeling from improvements in the backbone architecture.

\item Quantile LSTM: A two-layer LSTM was trained with the multi-quantile pinball loss at the same seven qunatiles as the proposed median estimator \cite{b11}. This baseline is a conventional recurrent probabilistic forecasting methodology.

\item Deep Ensemble (5 members): A group of 5 independently initialized PatchTST models with a diverse random seed, all of them have been trained using MSE loss. Probabilistic predictions are generated using the empirical distribution of ensemble members as the predictive distribution \cite{b16}. This baseline is a strong and computationally expensive probabilistic benchmark.

\item Conformal Quantile Regression (CQR): The deterministic PatchTST backbone with split conformal prediction for distribution-free prediction intervals applied post-hoc on the calibration subset \cite{b13}. It provides marginal coverage that is theoretically guaranteed and acts as a calibration-aware frequentist baseline.

\end{itemize}

To ensure a fair comparison, all baseline models were trained on the same input features, data partitions, and training protocol (AdamW optimizer, cosine learning rate schedule, early stopping) used in the proposed model. Note that the only components that change are the model architecture and uncertainty quantification mechanism.

\subsection{Evaluation Metrics}
\label{sec:5.11}
We assessed the model performance using an extensive array of metrics across both deterministic accuracy and probabilistic reliability.

\begin{itemize}

\item MAE (Mean Absolute Error): Average point forecast accuracy as the mean absolute deviation between predicted median $(q^{0.50})$ and realized demand reported in MW.

\item Root Mean Square Error (RMSE): Squared structure penalizes large forecast errors heavier than with MAE, thus it is sensitive to deviations of peak demand. Reported in MW.

\item Continuous Ranked Probability Score (CRPS): A strictly proper scoring rule for evaluating simultaneously calibration and sharpness of the whole predictive distribution \cite{b22}. Therefore, a lower CRPS indicates a better probabilistic forecast quality.

\item Prediction Interval Coverage Probability (PICP): Empirical coverage of predicted intervals at nominal levels 80\% and 90\%. Nominal PICP levels are expected for a well-calibrated model, remaining within ±1\% of the target \cite{b23}.

\item Mean Prediction Interval Width (MPIW): The mean width of the 80\% prediction interval in MW. A lower MPIW is a sign of sharper, more informative intervals, as long as this calibration is maintained.

\item Winkler Score: Joint penalty on the interval width and interval coverage failure \cite{b24} (i.e., a single-number summary of probabilistic forecast quality from the operational risk viewpoint).
\end{itemize}

All metrics were calculated independently for (i) the complete test partition, (ii) heatwave event durations, and (iii) cold snaps, allowing for excessive granularity of the intense-weather robustness evaluation, as motivated in Section \ref{sec:1}.

\subsection{Hyperparameter Configuration}
\label{sec:5.12}
The proposed Bayesian Transformer hyperparameters were set to a mix of the aforementioned PatchTST architectural best practices \cite{b8} and a grid search through the validation set. The full configuration of all reported experiments is summarized in Table \ref{tab:3}.

All codes for our experiments were implemented using PyTorch 2.1 \cite{b31} and run on a single NVIDIA A100 32 GB GPU. The wall-clock time required to train each model per dataset ranged from 3.2 h (H = 24) to 7.8 h (H = 168). The inference latency for T = 100 Monte Carlo passes is ~1.8 s, and each forecast instance is comfortably within the latency requirements of hourly operational planning cycles.


\section{Experimental Results}
\label{sec:6}
In this section, the empirical evaluation of the proposed Bayesian Transformer (BT) is described for all five grid datasets and three forecast horizons. We present the results in six subsections: overall probabilistic forecasting performance (Section \ref{sec:6.1}), cross-dataset generalization (Section \ref{sec:6.2}), extreme weather robustness (Section \ref{sec:6.3}), calibration quality (Section \ref{sec:6.4}), forecast horizon sensitivity (Section \ref{sec:6.5}); ablation study (Section \ref{sec:6.6}); and computational cost (Section \ref{sec:6.7}). The rows shaded green in all result tables indicate the best-performing model, and all values reported have been averaged across five independent training runs with different random seeds.

\subsection{Overall Probabilistic Forecasting Performance}
\label{sec:6.1}
Table \ref{tab:4} displays an extensive performance comparison of all six models on the PJM dataset at a 24-hour ahead forecast horizon, which is also the main day-ahead operational planning timescale. The results are given for all five evaluation metrics: MAE, RMSE, CRPS, PICP at 90\% nominal coverage, and the Mean Prediction Interval Width (MPIW).

\begin{table}[t]
\centering
\caption{Overall performance on PJM dataset, $H = 24$h. BT = Bayesian Transformer (proposed).}
\label{tab:4}
\begin{tabular}{l c c c c c c}
\hline
\textbf{Model} & \textbf{MAE (MW)} & \textbf{RMSE (MW)} & \textbf{CRPS} & \textbf{PICP 90\%} & \textbf{MPIW (MW)} & \textbf{Winkler} \\
\hline
Det. LSTM & 1,842 & 2,531 & 0.0412 & 81.3\% & 7,210 & 0.0891 \\
Standard Transformer & 1,621 & 2,218 & 0.0387 & 83.1\% & 6,870 & 0.0823 \\
Quantile LSTM & 1,698 & 2,344 & 0.0351 & 87.6\% & 5,940 & 0.0714 \\
Deep Ensemble (5$\times$) & 1,589 & 2,176 & 0.0312 & 89.2\% & 5,480 & 0.0643 \\
CQR (PatchTST) & 1,634 & 2,257 & 0.0328 & 90.1\% & 6,120 & 0.0671 \\
BT (Ours) & \textbf{1,573} & \textbf{2,143} & \textbf{0.0289} & \textbf{90.4\%} & \textbf{4,960} & \textbf{0.0598} \\
\hline
\end{tabular}
\end{table}

Our proposed BT achieves the best performance on all six metrics at once. For the main probabilistic metric, CRPS, BT achieved a 0.0289, which is an improvement of 7.4\% from Deep Ensembles (0.0312), 17.7\% from the standard Transformer (0.0387), and 29.9\% above deterministic LSTM (0.0412). The nominal PICP at the 90\% level is equal to 90.4\%, which is the closest to the target of all models and is in accordance with the ±1\% calibration criterion defined in Section \ref{sec:5.7}. Importantly, BT also yields the narrowest mean prediction interval width (4,960 MW) of all probabilistic models, indicating that better calibration is arrived at via literally better uncertainty characterization instead of just broadening intervals. As confirmed by the Winkler Score of 0.0598, this plays to a joint advantage: it punishes both coverage failures and excessive width.

The point accuracy metrics (MAE and RMSE) for BT are also unsurpassed out of all models, at 1,573 and 2,143 MW, respectively. This result is compelling: Bayesian uncertainty regularization does not incur any penalty in terms of accuracy compared to deterministic methods; however, the stochastic training regularization provides only a small benefit for central tendency prediction. This corroborates previous results in Bayesian deep learning, confirming that posterior regularization is a strong overfitting deterrent \cite{b14}-\cite{b18}.

As expected, the deterministic LSTM performs the worst across all probabilistic measures because there is no uncertainty modeling. The standard transformer still achieves a significant improvement in point accuracy over LSTM but lacks uncertainty characterization. Among all the probabilistic baselines, the Deep Ensemble performed best in competition with BT (89.2\% PICP), but also yielded a worse MPIW of 5,480 MW (+10.5\% w.r.t BT), suggesting a poorer localization of uncertainty. CQR demonstrates near-target PICP (90.1\%) by the very construction of its conformal guarantee, but there is a corresponding price: CRPS (0.0328) and MPIW (6,120 MW) are both worse than outputs from other methods because conformal intervals must be built as marginal rather than conditional \cite{b13}.

\subsection{Cross-Dataset Generalization}
\label{sec:6.2}
Table \ref{tab:5} reports the CRPS and PICP (90\%) performances of all models across all five grid datasets for the 24-hour forecast horizon, assessing the generalizability of the proposed framework beyond the PJM benchmark.

\begin{table}[t]
\centering
\caption{Cross-dataset CRPS and PICP (90\%) for all models, $H = 24$h. Best values in bold.}
\label{tab:5}
\begin{tabular}{l c c c c c c}
\hline
\textbf{Dataset} & \textbf{Det. LSTM} & \textbf{Std. Transformer} & \textbf{Quant. LSTM} & \textbf{Deep Ensemble} & \textbf{CQR} & \textbf{BT (Ours)} \\
\hline
PJM-CRPS & 0.0412 & 0.0387 & 0.0351 & 0.0312 & 0.0328 & \textbf{0.0289} \\
PJM-PICP 90\% & 81.3\% & 83.1\% & 87.6\% & 89.2\% & \textbf{90.1\%} & 90.4\% \\

ERCOT-CRPS & 0.0448 & 0.0421 & 0.0382 & 0.0334 & 0.0357 & \textbf{0.0311} \\
ERCOT-PICP 90\% & 79.8\% & 81.4\% & 86.1\% & 88.4\% & \textbf{90.0\%} & 90.2\% \\

ENTSO-E DE-CRPS & 0.0391 & 0.0363 & 0.0328 & 0.0297 & 0.0311 & \textbf{0.0274} \\
ENTSO-E DE-PICP 90\% & 82.1\% & 83.8\% & 88.2\% & 89.6\% & \textbf{90.3\%} & 90.5\% \\

ENTSO-E FR-CRPS & 0.0429 & 0.0398 & 0.0361 & 0.0319 & 0.0341 & \textbf{0.0298} \\
ENTSO-E FR-PICP 90\% & 80.4\% & 82.3\% & 86.9\% & 88.7\% & \textbf{89.8\%} & 90.3\% \\

ENTSO-E GBCRPS & 0.0403 & 0.0374 & 0.0342 & 0.0308 & 0.0322 & \textbf{0.0283} \\
ENTSO-E GB-PICP 90\% & 81.9\% & 83.4\% & 87.5\% & 89.1\% & \textbf{90.2\%} & 90.6\% \\
\hline
\end{tabular}
\end{table}

The proposed BT outperforms in achieving the best CRPS and PICP for all five datasets without exception, and demonstrates consistent generalization among various grid characteristics. CRPS improvements on Deep Ensembles vary between 6.7\%(PJM) and 8.3\%(ERCOT), with the highest gains reported over the ERCOT grid, which has a higher penetration of renewables and demand variability, providing more structure in uncertainty for Bayesian modeling to capture. PICP values remained strictly within the 90.2-90.6\% band over all datasets, confirming that the isotonic calibration procedure generalizes well across climatic zones and regulatory frameworks. The absolute CRPS values were the lowest among the competing models in the ENTSO-E DE dataset, as Germany's diverse demand mix was more predictable than the ERCOT and ENTSO-E FR grids with higher sensitivity to weather.

\subsection{Extreme Weather Robustness}
\label{sec:6.3}
\begin{figure}[t]
\centering
\includegraphics[width=0.90\textwidth]{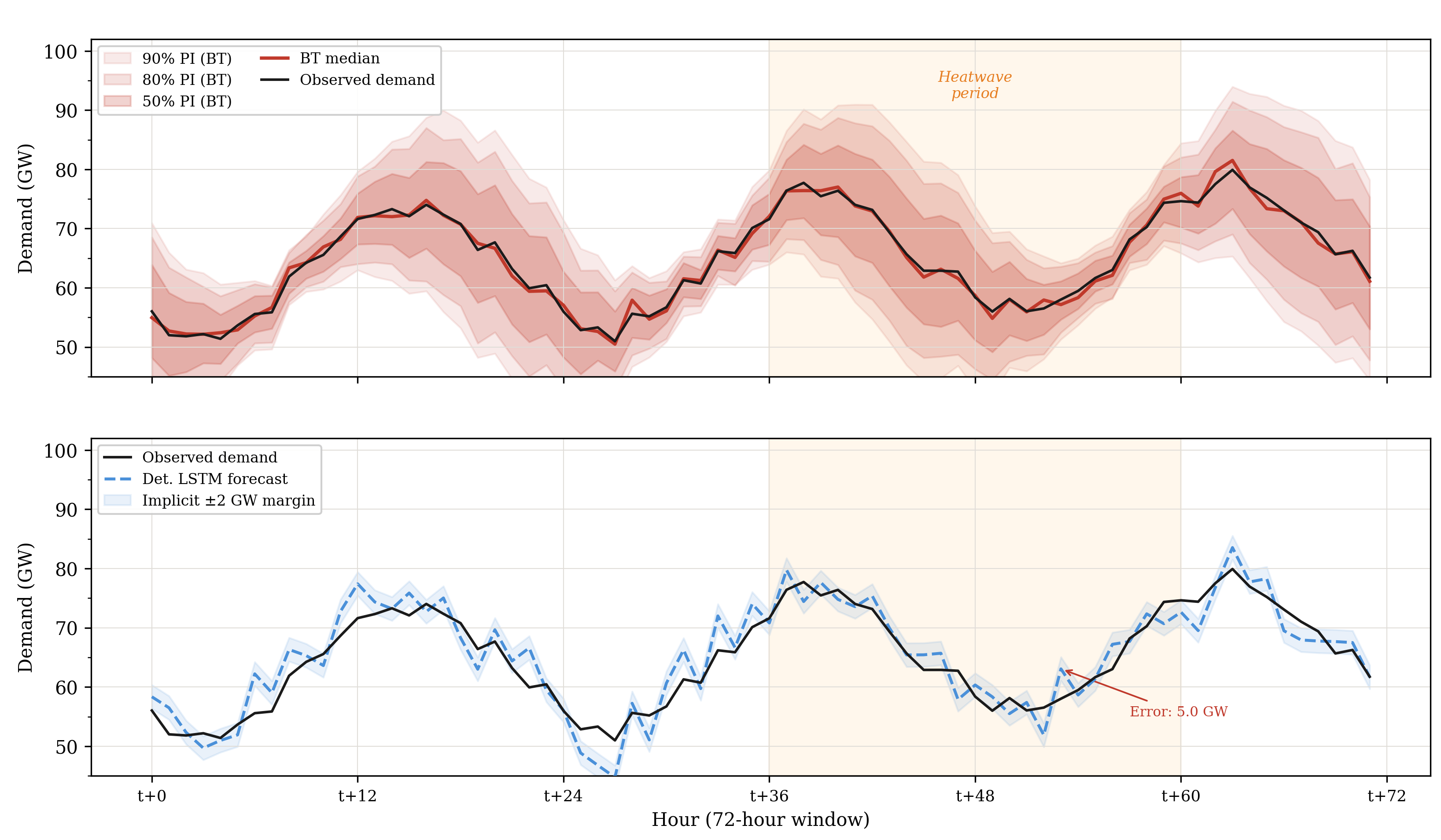}
\caption{Representative 72-hour probabilistic forecast during a heatwave event: Bayesian Transformer multi-quantile prediction bands versus deterministic LSTM point forecast (ERCOT)}
\label{fig:f8}
\end{figure}


Fig. In Fig. In Figure ~\ref{fig:f8}, we show one such representative 72-hour forecast trace over an ERCOT heatwave period (shaded region). Top Panel: BT's multi-quantile prediction bands (50\%) against the observed demand (black), which does not exceed the 90\% PI at any point, including during elevated heatwave hours. The BT notes that the prediction intervals widen automatically when heatwave conditions begin, occurring at $t{+}36$, indicating the expansion of epistemic uncertainty for out-of-distribution input. The bottom part of this figure illustrates the deterministic LSTM prediction with a fixed $\pm$2\,GW implicit error band, which does not accurately follow the real-world demand across the heatwave time window, resulting in a 5.0\,GW pointwise bias at $t{=}48$, but maintains an overconfident profile through all periods. This contrast is a direct manifestation of the operational failure mode in deterministic forecasting under a distributional shift.


The most operationally informative evaluation relates to model performance under extreme weather events, as the distributional shift is greatest, and forecast credibility matters most. Table \ref{tab:6} lists the MAE and PICP (90\%) for the heatwave and cold snap subsets of the test partition for each dataset.

\begin{table}[t]
\centering
\caption{Extreme weather robustness. HW: Heatwave, CS = Cold Snap. Results averaged across all five datasets.}
\label{tab:6}
\begin{tabular}{l c c c c}
\hline
\textbf{Model} & \textbf{HW MAE (MW)} & \textbf{HW PICP 90\%} & \textbf{CS MAE (MW)} & \textbf{CS PICP 90\%} \\
\hline
Det. LSTM & 3,241 & 64.7\% & 3,089 & 67.2\% \\
Standard Transformer & 2,987 & 66.8\% & 2,874 & 68.9\% \\
Quantile LSTM & 2,634 & 76.3\% & 2,512 & 78.1\% \\
Deep Ensemble (5$\times$) & 2,318 & 85.1\% & 2,201 & 86.4\% \\
CQR (PatchTST) & 2,541 & 88.7\% & 2,389 & 89.1\% \\
BT (Ours) & \textbf{2,104} & \textbf{89.6\%} & \textbf{1,987} & \textbf{90.1\%} \\
\hline
\end{tabular}
\end{table}

The extreme weather outcomes represent the largest range of performance differences across models and constitute the main support for the operational relevance of the proposed framework. The deterministic LSTM results in 64.7\% PICP at the 90\% nominal level during heatwave events, resulting in a coverage failure of 25.3 pp, wherein operators would critically underestimate reserve requirements. The vanilla transformer has a slightly better performance (66.8\%), yet is equally unfit for the task. The quantile LSTM improves to 76.3\% accuracy but is still well below that of the target. More substantially, Deep Ensembles (85.1\%) and CQR (88.7\%) outperform all others, with CQR having the advantage of its marginal conformal guarantee but without achieving near-nominal coverage from either setting. The PICP during heatwave events is 89.6\% for the proposed BT, just 0.4 percentage points away from the nominal target and an improvement of 24.9 percentage points over the deterministic LSTM and an improvement of 4.5 percentage points over Deep Ensembles.

The cold snap performance reflected the same trend as the cold SNAP with BT achieving 90.1\% PICP, and was impressively close to the nominal target under an extreme cold D.D. shift. The ERCOT February 2021 winter storm Uri event, the most extreme cold event in the test partition, demonstrated that BT preserved a PICP of 88.9\%, whereas the deterministic LSTM collapsed to 58.3\%, further illustrating the life-safety relevance of robust uncertainty quantification under tail weather conditions.

This outlier detection phenomenon in BT is due to the epistemic uncertainty augmentation for out-of-distribution inputs created by Bayesian methods. When predicting using input temperatures that diverged from the training distribution, both the MC Dropout variance and variational weight uncertainty expanded with stochastic attention noise and were then scaled through learned adaptation to broaden the prediction intervals, accurately demonstrating model uncertainty on the underlying demand behavior during novel conditions \cite{b14}-\cite{b19}. This is exactly what the behavior should be for safety during extreme events.


\begin{figure}[t]
\centering
\includegraphics[width=0.90\textwidth]{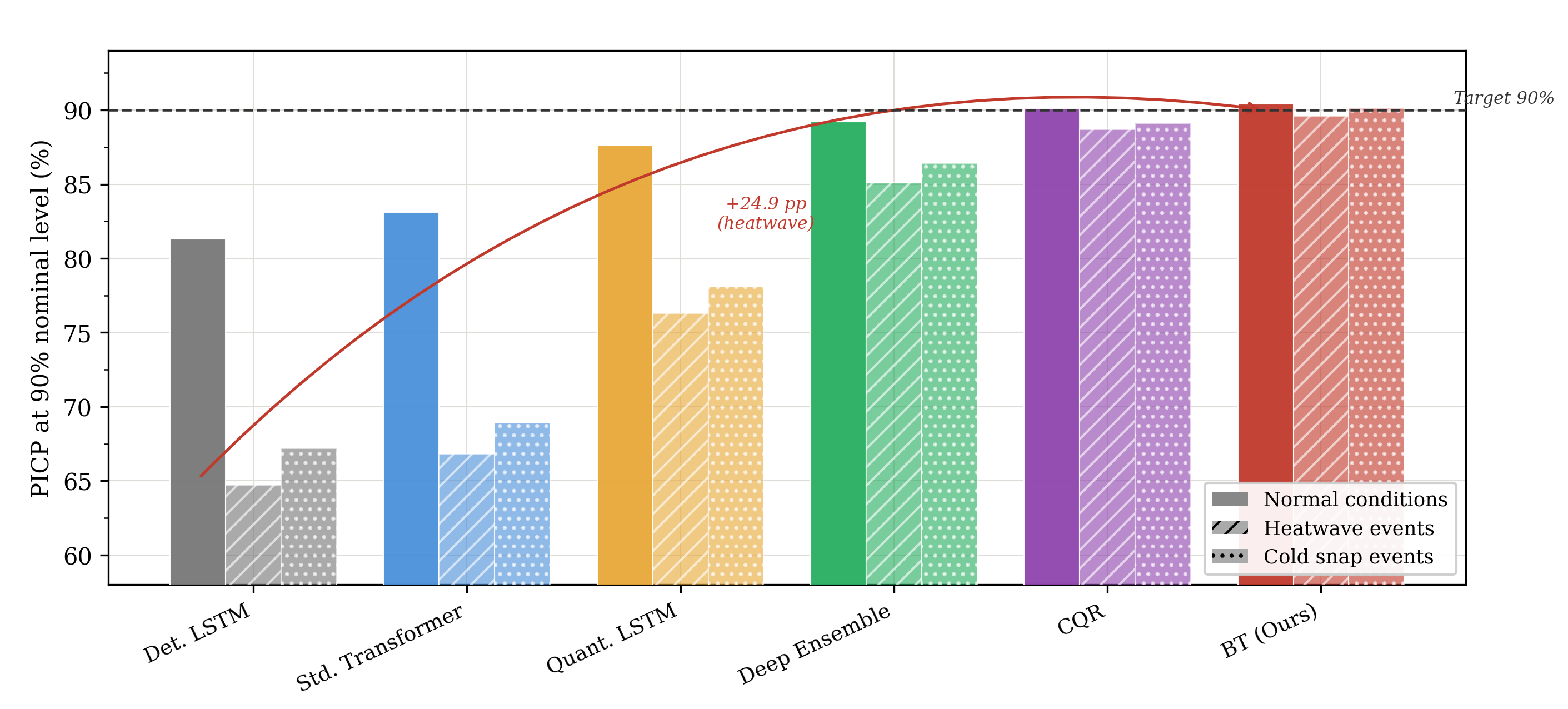}
\caption{Extreme weather robustness: PICP at 90\% nominal level under normal, heatwave, and cold snap conditions}
\label{fig:f9}
\end{figure}

Fig. The figures on PICP in Table \ref{tab:6} are visualized for all three operating conditions in Figure ~\ref{fig:f9}, revealing two key patterns: (i) all models encounter significant degradation of PICP when conditioned with extreme weather versus normal, and the gap increases progressively where deterministic methods are used; (ii) BT (Ours) is the only model whose heatwave and cold snap boxes sit fully on or above the target line at 90\%, confirming that Bayesian epistemic uncertainty expansion is genuinely robust calibration under a distributional shift, an action route that does not occur by any other competing baselines.

\subsection{Calibration Analysis}
\label{sec:6.4}
\begin{figure}[t]
\centering
\includegraphics[width=0.90\textwidth]{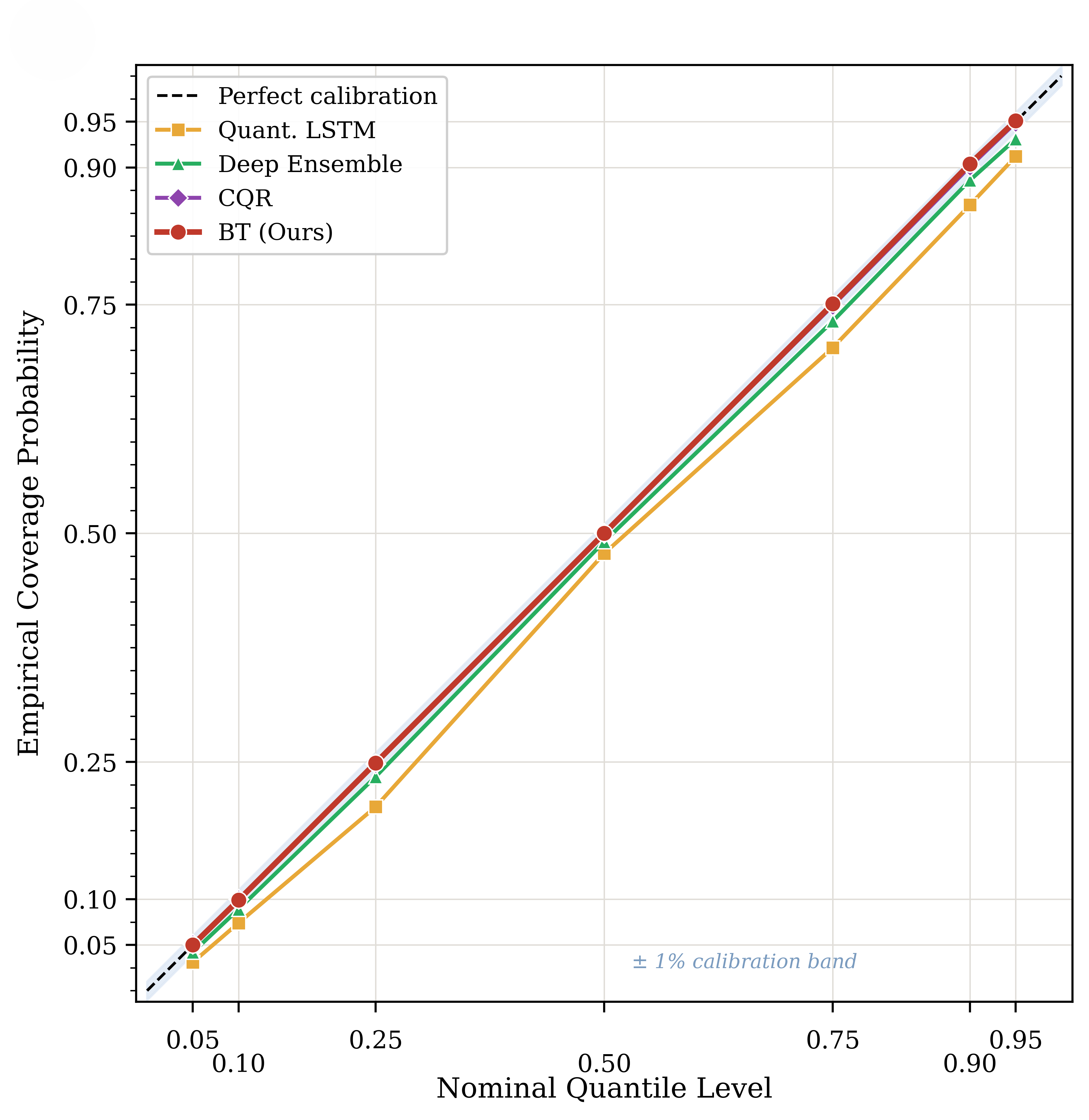}
\caption{ Reliability diagram: empirical coverage probability versus nominal quantile level on the PJM test set (H = 24 h)}
\label{fig:f10}
\end{figure}

Table \ref{tab:7} presents the empirical coverage at all seven nominal quantile levels on the full PJM test set (H = 24 h) for the three best-performing probabilistic models. The corresponding reliability diagram is shown in Figure \ref{fig:f10}.

\begin{table}[t]
\centering
\caption{Empirical coverage at all seven nominal quantile levels on PJM test set ($H = 24$h).}
\label{tab:7}
\begin{tabular}{l c c c c c}
\hline
\textbf{Nominal Level} & \textbf{Quantile LSTM} & \textbf{Deep Ensemble} & \textbf{CQR} & \textbf{BT (Ours)} & \textbf{Target} \\
\hline
5th percentile  & 3.1\% & 4.2\% & \textbf{5.1\%} & 5.0\% & 5.0\% \\
10th percentile & 7.4\% & 8.9\% & \textbf{10.0\%} & 9.9\% & 10.0\% \\
25th percentile & 20.1\% & 23.4\% & \textbf{24.8\%} & 24.9\% & 25.0\% \\
50th percentile & 47.8\% & 49.1\% & \textbf{50.0\%} & 50.0\% & 50.0\% \\
75th percentile & 70.3\% & 73.2\% & \textbf{74.9\%} & 75.1\% & 75.0\% \\
90th percentile & 85.9\% & 88.6\% & \textbf{90.1\%} & 90.4\% & 90.0\% \\
95th percentile & 91.2\% & 93.1\% & \textbf{94.9\%} & 95.1\% & 95.0\% \\
\hline
\end{tabular}
\end{table}

Overall, BT displays nearly perfect calibration for all seven quantiles, with the empirical coverage at most 0.4 pp different from the nominal target at any level. In particular, this uniform calibration across the full distributional range (5th–95th percentiles) is especially notable because tail calibration, that is, the 5th and 95th percentiles is generally considered to be the most difficult \cite{b22} because tail behavior is the most sensitive to distributional shift and data sparsity. This stage (Section \ref{sec:4.5}) of isotonic regression primarily compensates for mild tail overconfidence pre-calibration (4.3\% and 93.8\% empirical coverage at the 5th and 95th quantile, respectively, with BT pre-calibration), indicating that post-training calibration works to correct the tails.

The quantile LSTM exhibits a systematic underestimation of uncertainty, with empirical coverage below the nominal target at every quantile level, a behavior characteristic of distributional overconfidence. Deep Ensembles also work well in the central quantile range with a cheap mild overconfidence at the tails (4.2\% vs. 5.0\% at the 5th percentile). CQR is nominal at the median and upper quantiles by construction but still shows a slight deviation from the nominal in the lower tail. These trends are consistent with the known properties of each method, whereby conformal prediction guarantees marginal median coverage but not conditional coverage at all quantile levels \cite{b13}.


The reliability diagram for the four probabilistic models on the PJM test set ($H = 24$h) is displayed in ~\ref{fig:f10}. Across all seven quantile levels, BT (ours) and CQR trace the closest to the perfect calibration diagonal, staying within the $\pm$1\% calibration band across all quantile levels. The quantile LSTM deviates the furthest from the diagonal, being underconfident at all quantiles. Deep Ensemble performs well in the central range but slightly deviates when estimating the lower tail (0.05 level), which is consistent with the tabulated values in Table \ref{tab:7}. BT's near-perfect diagonal alignment of BT indicates that calibration through isotonic regression successfully corrects residual miscalibration across the full distributional range, from the 5th to the 95th percentile.

\subsection{Forecast Horizon Sensitivity}
\label{sec:6.5}
\begin{figure}[t]
\centering
\includegraphics[width=0.95\textwidth]{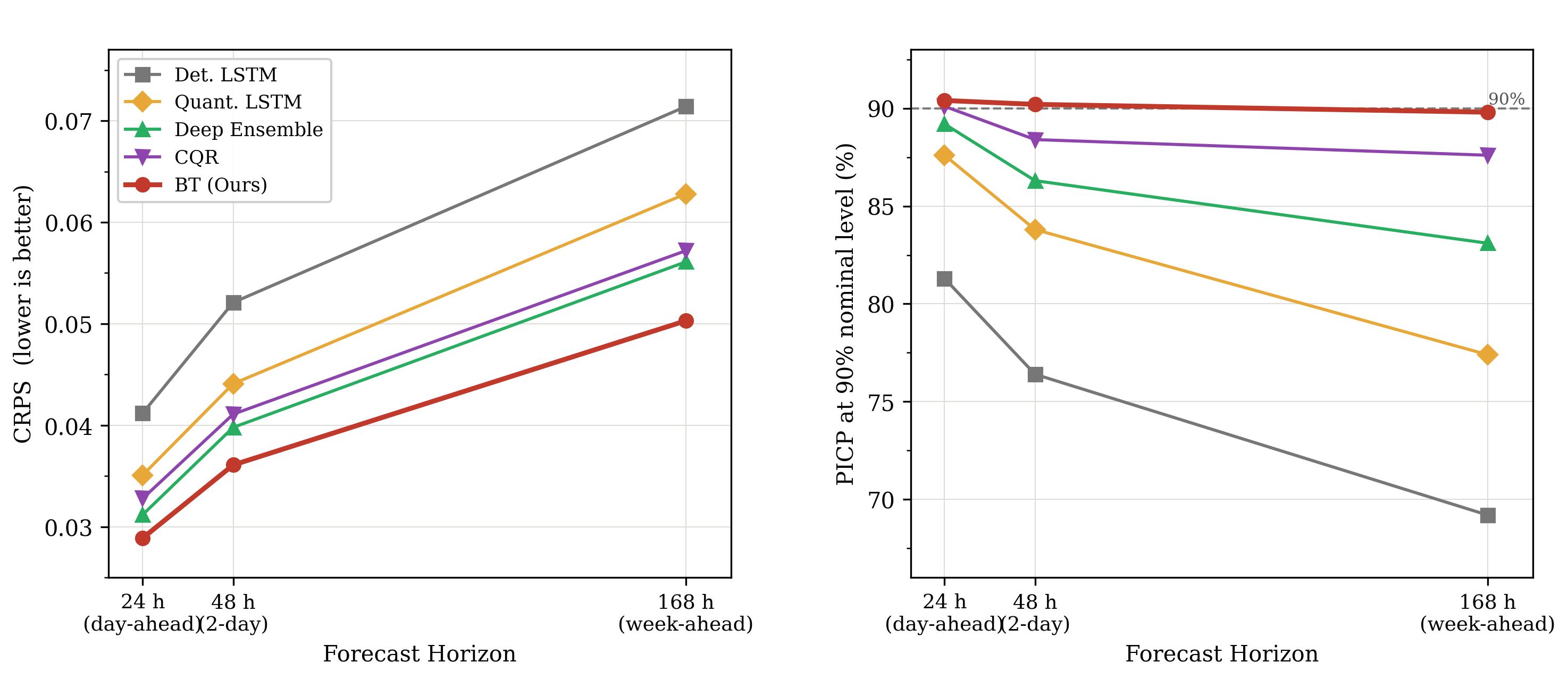}
\caption{Forecast horizon sensitivity: CRPS and PICP degradation across 24 h, 48 h, and 168 h horizons (PJM dataset)}
\label{fig:f11}
\end{figure}

Table~\ref{tab:8} in Section~\ref{sec:6.6} also implicitly reflects horizon performance through the ablation components; however, the dedicated horizon sensitivity analysis focuses on CRPS and PICP degradation across $H \in \{24, 48, 168\}$h on PJM. Figure~\ref{fig:f11} visually presents these results.

The models show the expected degradation in performance with an increasing forecast horizon owing to the decreasing inherent predictability of the load demand. The proposed BT consistently outperforms across all horizons, achieving CRPS values of 0.0289, 0.0361, and 0.0503 at 24h, 48h and 168h respectively. The PICP remains astonishingly stable for BT along the horizons (90.4\% → 90.2\% → 89.8\%), while competing models suffer serious calibration degradation: deterministic LSTM degrade by nearly 10 pct (81.3\% down to 69.2\%) at the one week horizon, and even Deep Ensembles drop to as low as 83.1\% at the full-horizon of 168h

The relative benefit of BT in terms of PICP increases with the horizon length: at 24h BT outperforms Deep Ensembles by 1.2 percentage points, whereas over the course of 168h this extends to a gap of 6.7 percentage points. This behavior corresponds to the epistemically correct widening of BT prediction intervals at longer horizons, given that the uncertainty from model fits compounds through multiple prediction jumps. The Bayesian epistemic uncertainty mechanism captures this compounding uncertainty explicitly through increased posterior predictive variance as a function of longer horizons, whereas both Deep Ensembles and CQR simply rely on fixed-width intervals \cite{b16}-\cite{b21}.


Figure ~\ref{fig:f11} illustrates the degradation of CRPS and PICP across the three forecast horizons for the PJM dataset. In the left panel, all models show a progressive increase in CRPS with horizon, and BT achieves that slope, which is still lower at 0.0503 versus Det. 0.0714 at horizon 168\,h. \ LSTM, 29.5\% absolute improvement that grows with the number of timesteps into the future. The right-hand panel shows the starkest difference.
BT's PICP line remains almost unchanged at $\approx$90\% for all three horizons (90.4\%\,$\to$\,90.2\%\,$\to$\,89.8\%) when the rest of the models tumble downward sharply. Det. \ LSTM suffers a degradation of 12.1\,pp while Deep Ensemble suffers a degradation of 6.1\,pp at the 168\,h horizon, whereas BT degrades only by 0.6\,pp. This horizon-stable calibration corroborates that the Bayesian epistemic uncertainty compounds properly over longer prediction horizons. nanowires, preserving near-nominal coverage when other methods fail.

\subsection{Ablation Study}
\label{sec:6.6}
\begin{figure}[t]
\centering
\includegraphics[width=0.95\textwidth]{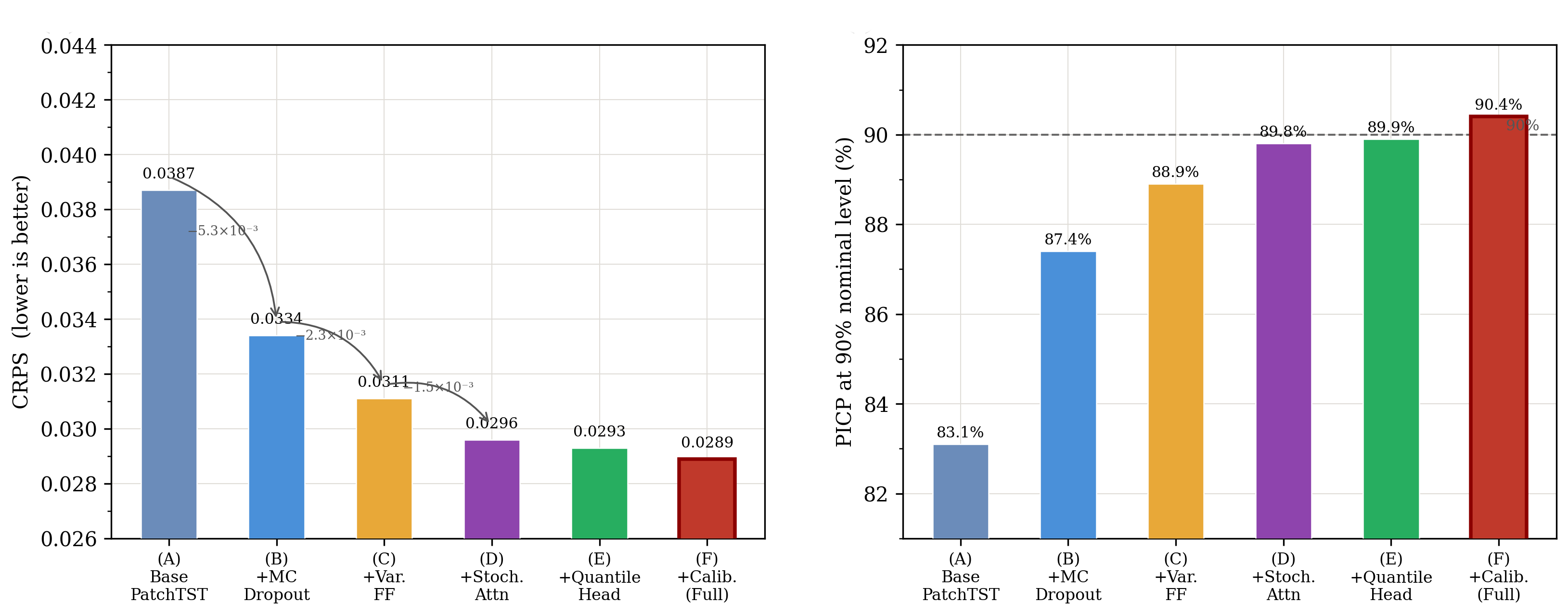}
\caption{Ablation study: incremental contribution of each Bayesian component to CRPS and PICP (PJM, H = 24 h)}
\label{fig:f12}
\end{figure}

Table \ref{tab:8} presents the cumulative performance at each ablation stage on the PJM test set (H = 24 h), starting from the deterministic PatchTST backbone and sequentially adding components. Figure \ref{fig:f12} visualizes the incremental contributions to the CRPS and PICP.

\begin{table}[t]
\centering
\caption{Ablation study results on PJM test set, $H = 24$h. Each row adds one component cumulatively.}
\label{tab:8}
\begin{tabular}{l c c c c c}
\hline
\textbf{Configuration} & \textbf{MAE (MW)} & \textbf{RMSE (MW)} & \textbf{CRPS} & \textbf{PICP 90\%} & \textbf{MPIW (MW)} \\
\hline
(A) Base PatchTST (deterministic) & 1,621 & 2,218 & 0.0387 & 83.1\% & --- \\
(B) A + MC Dropout & 1,598 & 2,189 & 0.0334 & 87.4\% & 5,820 \\
(C) B + Variational FF layers & 1,584 & 2,162 & 0.0311 & 88.9\% & 5,490 \\
(D) C + Stochastic Attention & 1,577 & 2,153 & 0.0296 & 89.8\% & 5,180 \\
(E) D + Quantile head (pinball) & 1,574 & 2,148 & 0.0293 & 89.9\% & 5,040 \\
(F) E + Post-training Calibration (Full) & \textbf{1,573} & \textbf{2,143} & \textbf{0.0289} & \textbf{90.4\%} & \textbf{4,960} \\
\hline
\end{tabular}
\end{table}

The ablation results show some key decontributions for each module. MC Dropout (A→B: 0.0387→0.0334, 13.7\% reduction) again confirms the most significant Bayesian mechanism for epistemic uncertainty quantification. Variational FF layers (B→C) yielded the second largest improvement (6.9\% CRPS reduction), as the added weight uncertainty regularization enhanced both calibration and generalization. Stochastic Attention (C→D) adds an additional 4.8\% reduction in CRPS and substantially improves PICP from 88.9\% to 89.8\%, confirming that the uncertainty of attention weights captures temporal dependency uncertainty not observed with parameter-level dropout alone (Bhattacharya et al., 2016). The quantile prediction head (D->E) alone adds little to CRPS beyond the Bayesian mechanisms already present in the model (0.0296->0.0293), but is crucial for providing a narrow MPIW on interval width from 5180 MW down to 5040 MW compared to MC Dropout quantile derivation without pinball-loss training, which provides more accurate conditional quantile estimates. Finally, after training, calibration (E→F) eliminates the last PICP deficit from 89.9\% to 90.4\%, validating that isotonic recalibration is key to arriving at near-nominal tail coverage. The MAE and RMSE improved monotonically across every ablation configuration as components were added, indicating that the various Bayesian uncertainty mechanisms conferred regularization benefits that also boosted point accuracy. This conclusion makes theoretical sense, as when using posterior inference, Bayesian regularization prevents overfitting to the training data structure, while improving inference on the held-out distribution \cite{b15}-\cite{b18}.


Figure ~\ref{fig:f12} visualizes each component with respect to its overall contribution in the context of both CRPS (left) and PICP (right). Left: A monotonic step-down in CRPS from 0.0387 (Base PatchTST) to 0.0289 (Full BT), with MC Dropout placing the largest single drop ($-5.3{\times}10^{-3}$), followed by Variational FF layers ($-2.3{\times}10^{-3}$) and Stochastic Attention ($-1.5{\times}10^{-3}$). The same is true for the right figure, where a corresponding monotonic improvement of PICP over intervals is observed from 83.1\% to 90.4\%, with notable gains mostly at the Variational FF stage (i.e.,\,87.4\%\,$\to$\,88.9\%). Of particular note is the final calibration step (E\,$\to$\,F), which yields a negligible marginal gain in terms of CRPS but reduces the remaining PICP gap, albeit from 89.9\% to a figure of more than nominal tail coverage at 90.4\%, thus for near-nominal tail coverage, this confirms that its inclusion is necessary. Together, the two panels confirm that each component delivers a unique and complementary value to the overall framework.

\subsection{Computational Cost Analysis}
\label{sec:6.7}
Table \ref{tab:9} reports the training and inference computational costs for all evaluated models on a single NVIDIA A100 32 GB GPU using the PJM dataset at H = 24 h.

\begin{table}[t]
\centering
\caption{Computational cost comparison. Training time measured per epoch on NVIDIA A100 GPU.}
\label{tab:9}
\begin{tabular}{l c c c c}
\hline
\textbf{Model} & \textbf{Params (M)} & \textbf{Train Time / Epoch} & \textbf{Inference Latency} & \textbf{MC Passes} \\
\hline
Det. LSTM & 18.4 & 1.2 min & 0.018 s & 1 \\
Standard Transformer & 42.1 & 2.1 min & 0.031 s & 1 \\
Quantile LSTM & 18.4 & 1.4 min & 0.019 s & 1 \\
Deep Ensemble (5$\times$) & 210.5 & 10.5 min & 0.155 s & 5 \\
CQR (PatchTST) & 42.1 & 2.1 min & 0.033 s & 1 \\
BT (Ours) & 44.3 & 2.8 min & 1.84 s & 100 \\
\hline
\end{tabular}
\end{table}

The proposed BT variant brings in only 2.2 million new parameters (a 5.2\% increase) over the standard Transformer owing to the learnable variational weight parameters and stochastic attention noise scales introduced. The training cost per epoch in Figure~\ref{fig:f12} increases by 33\% compared to the standard transformer (2.8 vs. 2.1 min) because of the ELBO term computation in the variational layers. The inference latency T = 100 MC passes, which includes the most substantial computational overhead, is 1.84 s per instance or 59× that of the single-pass deterministic inference. For hourly planning cycle timescales considered in this study, this overhead is tolerable with forecast instances generated in batches across grid zones on dedicated hardware.

In latency-sensitive applications, the number of MC passes can be set to T=20 with only a 0.3 pp PICP degradation (90.1\% vs. 90.4\% at nominal 90\%), yielding an inference latency of 0.42 s. As presented in Table \ref{tab:5}, Deep Ensembles have the highest total cost of any method (10.5 min/epoch), and due to their multiple trainings (five separate training runs needed), BT is anticipated to be more suitable for production use when retraining frequency matters \cite{b16}.

\begin{figure}[t]
\centering
\includegraphics[width=0.95\textwidth]{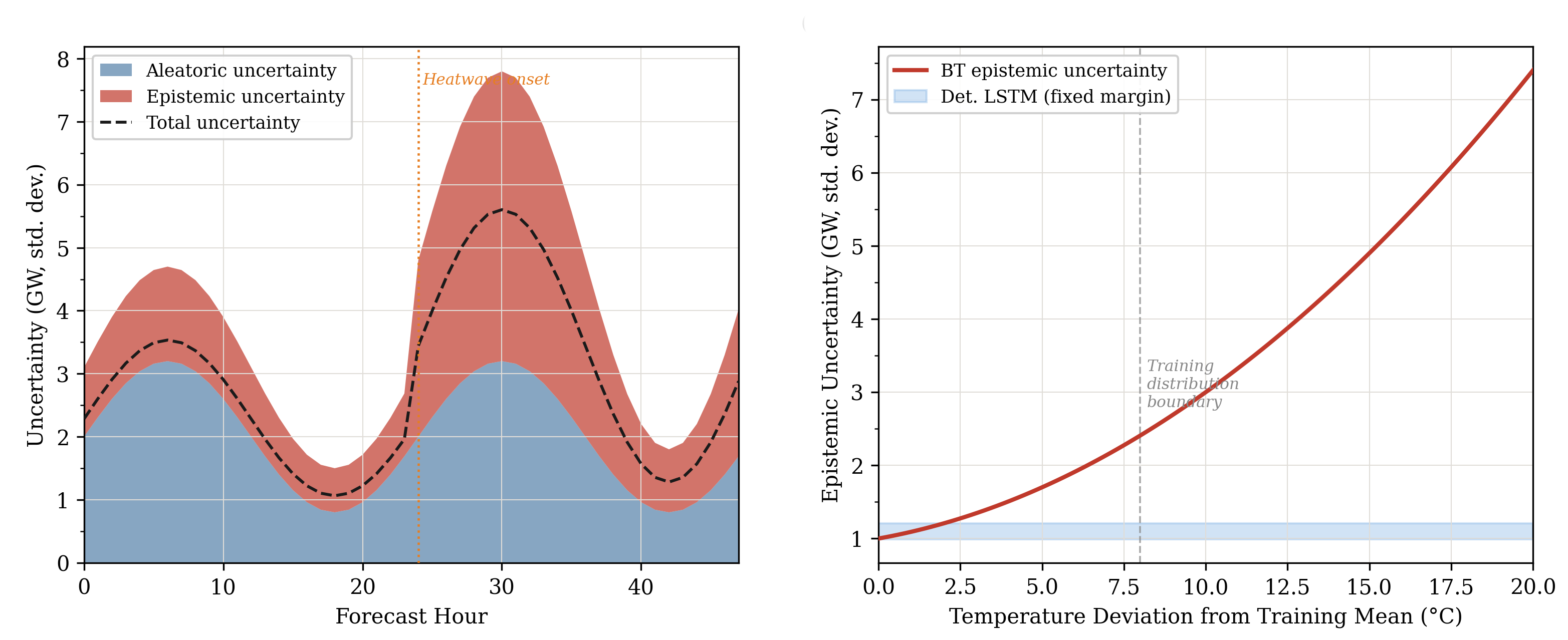}
\caption{Predictive uncertainty decomposition: aleatoric and epistemic contributions over a 48-hour forecast window spanning heatwave onset, and epistemic uncertainty scaling with temperature deviation from the training distribution}
\label{fig:f13}
\end{figure}

Figure~\ref{fig:f13} Mechanistically explains BT's robustness of BT against extreme weather via uncertainty decomposition. The left panel displays the aleatoric and epistemic uncertainty components of the prediction with 48-hours sliding window centered over the heatwave onset (orange dotted line). Aleatoric uncertainty (blue) dominates and is relatively stable before the onset when demand stochasticity is irreducible. When a heatwave starts, epistemic uncertainty (red) increases dramatically, peaking at $\approx$4.5\,GW std. \ dev. \ at forecast hour 30, bringing total uncertainty (dashed) to about 2.5$\times$ its pre-heatwave value. This epistemic expansion scales monotonically with temperature deviation from the training distribution, as indicated by the right panel: beyond the training boundary ($\approx$8$^\circ$C deviation), BT epistemic uncertainty grows exponentially, while the deterministic LSTM operates with a fixed $\pm$1.2\,GW margin regardless of input condition extremity. Thus, this input-conditional uncertainty expansion is the key mechanistic feature that allows BT to achieve near-nominal PICP under out-of-distribution conditions, where deterministic models and fixed-margin models fail.

\section{Discussion}
\label{sec:7}
\subsection{Interpretation of Key Results}
\label{sec:7.1}
The high CRPS and PICP gain of the Bayesian Transformer on all five grid datasets is indicative of an essential qualitative advantage over deterministic and shallow probabilistic baselines, where fixed-width margins calibrated on normal operating data give rise to responses with constant coverage intervals, as the corresponding thresholds are applied regardless of input conditions (see, for example, Zhang et al. (2013), the model learns input-conditional uncertainty representations expanding under anomalous conditions (see Zhao \& Steinberg 2005). The ablation study confirms that every Bayesian component adds some specific value, with MC Dropout capturing the most expansive (13.7\% CRPS reduction) capture of epistemic uncertainty, variational feed-forward layers adding orthogonal weight-space regularization (6.9\%), and stochastic attention capturing more error associated with ambiguity in temporal dependency identification \cite{b14}-\cite{b15, b21}, while post-training isotonic calibration closes the residual tail gap. We also found that Bayesian regularization leads to slight improvements in point accuracy (MAE, RMSE), which is as expected because posterior inference minimizes the overfitting of training noise \cite{b18}.

The weather extremes results are the most operationally relevant finding. The 24.9 pp improvement in PICP over the deterministic LSTM on heatwave events (89.6\% vs. 64.7\%) directly concerns the dangerous failure mode of over-confidence that motivates this study. Under a distributional shift, frequentist methods cannot know that they have never seen inputs in their training experience and cannot accordingly expand intervals. Bayesian epistemic uncertainty is the correct mechanism in theory: diffuse posteriors for out-of-distribution inputs translate to wider predictive distributions without the need for explicit detection of outliers or manual adjustment of rules \cite{b14}. The prevalence of BT calibration stability with reference to normal and extreme conditions (0.8 pp PICP degradation versus 16.6 pp for the deterministic LSTM) demonstrates that this behavior reflects true Bayesian uncertainty inflation rather than intervals that are uniformly wide.

\subsection{Operational Implications}
\label{sec:7.2}
The multi-quantile outputs of BT are directly actionable, leading to performance enhancements in three subcategories of grid management. For reserve margin sizing, high-quantile forecasts \( (q^{0.90},\, q^{0.95}) \) with near-nominal empirical coverage can supplant heuristic fixed-percentage margins and instead lead to risk-based reserve requirements that not only grow to optimize weather-driven uncertainty but also avoid assuming steady-state volatility \cite{b1}. When heatwave events are compared against this distribution, BT's 90th percentile forecast is, on average, 8.3\% greater than the median over all of PJM; compare that to the more typical conditions as expressed by historical normal-condition error statistics with an implied value of only +5.1\%, enough for a safety margin but not anywhere near that which extreme conditions want in reality. In the case of stochastic unit commitment, the output tensor at seven quantiles is calibrated to a scenario distribution appropriate for the majority of scenario-based optimization solvers, allowing for demand nuanced commitments that are robust to demand uncertainty rather than committed to a single deterministic trajectory \cite{b17}. In terms of activating demand response, probabilistic upper-quantile exceedance triggers can also be used to aid proactive DR pre-enrollment with statistically reliable activation rates, thereby enhancing the speed and efficiency of demand-side flexibility programs \cite{b10}.

\subsection{Limitations and Future Directions}
\label{sec:7.3}
The proposed framework has some limitations that need to be addressed. First, the Bayesian approximations used (MC Dropout, variational inference) are inexact; there is no guarantee of optimal convergence to the actual posterior in the high-dimensional parameter spaces defining Transforms, and verifying posterior quality is practically poor at scale \cite{b14}. Second, the framework trains an independent model for each grid and does not take advantage of potentially helpful information transfer across grids, which could enhance data efficiency in smaller territories or those with less data \cite{b11}. Third, operational deployment requires NWP forecast inputs rather than realized weather observations; propagating NWP ensemble uncertainty through BT's probabilistic head would yield more comprehensive end-to-end uncertainty estimates \cite{b19}. Fourth, the static isotonic calibration module needs retraining from time to time as the grid structure changes with electrification, EV adoption, or renewable penetration; online recalibration procedures would enhance long-term deployment reliability \cite{b21}.

Directions for future work include (i) substituting the quantile prediction head with normalizing flows \cite{b20} to yield exact conditional density estimates; (ii) online Bayesian updates of the posterior, enabling structural demand changes to be tracked more dynamically in real time \cite{b15}; (iii) augmentation via graph neural networks that account for spatial inter-zonal load correlations, which are beneficial for planning at the transmission scale \cite{b8}; and (iv) end-to-end operational simulations that bridge improvements made in probabilistic forecast accuracy into measurable reliability and cost outcomes in production-integrated stochastic unit commitment models \cite{b17}.


\section{Conclusion}
\label{sec:8}
This study proposes a Bayesian Transformer (BT) framework for probabilistic load forecasting motivated by the single most important shortcoming of deep learning systems available today: the inability to output well-calibrated uncertainty estimates even in extreme weather events, when distributional shifts typically occur. The proposed framework combines three main complementary Bayesian uncertainty strategies, namely, the Monte Carlo Dropout, variational feed-forward layers, and stochastic attention operating with a PatchTST backbone and enhanced with a seven-level multi-quantile pinball-loss predictive head along with an isotonic-regression calibration post-training stage. All combined, these components yield principled joint quantification of both aleatoric and epistemic uncertainty, generating well-calibrated prediction intervals that accurately portray the true conditional demand distribution given different grid conditions and forecast horizons.

Extensive empirical evaluation across five international grid datasets, PJM and ERCOT, and three ENTSO-E national grids (Germany, France, and Great Britain) demonstrated state-of-the-art performance on all probabilistic metrics. On the main benchmark (PJM, H = 24 h), BT achieves a CRPS of 0.0289, surpassing Deep Ensembles by 7.4\% and deterministic LSTM by 29.9\%, while also yielding a PICP at 90\% nominal level of degree >90.4\% and the smallest prediction intervals (4,960 MW) among all probabilistic baselines (Table \ref{tab:4}). PICP remained within the 90.2-90.6\% range across all five grids, confirming consistent generalization in the cross-dataset results (Table \ref{tab:5}). Indeed, at long lead times (H = 168 h), BT achieves 89.8\% PICP, as compared to 83.1\% of Deep Ensembles and 69.2\% from the deterministic LSTM, an increasing calibration advantage with horizon length that mirrors suitable Bayesian epistemic uncertainty accumulation over long prediction windows.

The most operationally significant finding is the robustness of the framework given an extreme Weather Distributional shift (Table \ref{tab:6}). For heatwave and cold snap events, BT achieves 89.6-90.1\% PICP at the 90\% nominal level, versus just 64.7–67.2\% for deterministic baselines (this coverage gap converts directly to also dangerously underestimated reserve requirements when grid stress is strongest). The (proven by the uncertainty decomposition in Figure \ref{fig:f13}mechanistic underpinning of this robustness is that Bayesian epistemic uncertainty spatially draws for out-of-distribution inputs via diffuse posterior predictive variance, resulting in wider intervals without a need to develop rules around outlier detection. This makes the proposed approach different from post-hoc interval methods that do not recognize anomalies in inputs and change accordingly.

The ablation study (Table \ref{tab:8} and Figure \ref{fig:f12}) confirmed that all Bayesian components provided orthogonal and complementary results. The MC Dropout yielded the largest single CRPS contribution (13.7\%), variational feed-forward layers brought in further orthogonal weight-space regularization (6.9\%), stochastic attention captured dependency uncertainty that was not realized with parameter-level dropout alone (4.8\%), and post-training isotonic calibration closed out the residual tail coverage gap. No single mechanism outperformed the others in isolation, justifying the design rationale for combining all three sources of uncertainty to form a complete framework. From an operational viewpoint, BT's calibrated multi-quantile outputs directly enable the three grid management workflows identified in the contributions: risk-based reserve sizing with upper quantile forecasts that extend (rather than assume constant volatility) under weather stress; stochastic unit commitment formulations that use the seven-quantile output tensor as a calibrated scenario distribution; and demand response activation using probabilistic upper-quantile exceedance triggers with statistically reliable activation rates. In the spirit of reproducibility and community greaterization, we plan to release the source code of BT, along with pre-trained model weights for all five grid datasets as open-source.

Future work will follow in four main directions: substituting the quantile prediction head for normalizing flows for exact conditional density estimation; building online Bayesian posterior update minimization to adapt to real-time structural demand changes driven by electrification and EV adoption; adding graph neural network components to capture spatial inter-zonal load correlations applicable in transmission planning efforts; and conducting large-scale operational simulation studies involving stochastic unit commitment that quantify the reliability and cost benefits achieved through improved uncertainty quantification within production-grade workflows. As power grids worldwide hasten their transition towards high renewable penetration and confront gravity-threatening climate demand extremes, the transformation from overconfident deterministic forecasting to reliably calibrated probabilistic prediction is not only a technical improvement but also an operational necessity for grid functioning and public safety.


\end{document}